\documentclass[letterpaper,twocolumn,10pt]{article}
\usepackage{usenix-2020-09}

\usepackage{tikz}
\usetikzlibrary{positioning,shapes,arrows.meta,backgrounds,calc}
\usepackage{amsmath}
\usepackage{amssymb}
\usepackage{amsfonts}
\usepackage{booktabs}
\usepackage{hyperref}
\usepackage{graphicx}
\usepackage{subcaption}
\usepackage{pifont}
\usepackage{xcolor}
\graphicspath{{figures/}}
\usepackage{xspace}
\usepackage{enumitem}

\usepackage{filecontents}

\newcommand{\name}{{\texttt{ACA}}}
\newcommand{\nameLong}{Adversarial Calibration Attack}
\begin{document}


\title{\LARGE \bf Adversarial Calibration Attack on Autonomous Vehicles}

\author{
{\rm Liangkai Liu}\\
Texas Tech University
\and
{\rm Qingzhao Zhang}\\
University of Arizona
\and
{\rm Kang G. Shin}\\
University of Michigan
} 

\maketitle

\begin{abstract}

Autonomous vehicles (AVs) rely on accurate camera--LiDAR calibration for multimodal sensor fusion. In practice, calibration can drift due to vibration, temperature variation, or minor sensor displacement, motivating online calibration algorithms that detect and correct misalignment at runtime while allowing the vehicle to continue operating without a factory visit. Existing AV attacks largely assume correct calibration. We instead identify online sensor calibration as a new attack plane. A corrupted calibration update can persist across subsequent fusion operations, causing system-wide errors that propagate from perception to planning and control.

We present \emph{\nameLong} (\name), the first physical attack against camera--LiDAR online calibration. Using a single adversarial poster, \name\ first spoofs the miscalibration detector to trigger the calibration process and then steers the calibration estimator toward an incorrect transformation. A unified optimization jointly designs the poster's geometry and texture for both objectives. We evaluate \name\ across benchmark datasets, simulation, and physical experiments. On benchmark datasets such as KITTI and nuScenes, \name\ induces up to $33.9^\circ$ mean rotational calibration error, thereby severely degrading object detection. In the CARLA simulator, the attack causes a collision when the corrupted calibration is accepted in vulnerable scenarios crafted by the attacker. On a real Husky robot, a printed adversarial poster successfully reproduces the calibration error. These results demonstrate that online calibration is a practical and safety-critical attack surface for AVs.
\end{abstract}

\section{Introduction}
\begin{figure*}[t]
\centering
\includegraphics[width=.9\textwidth]{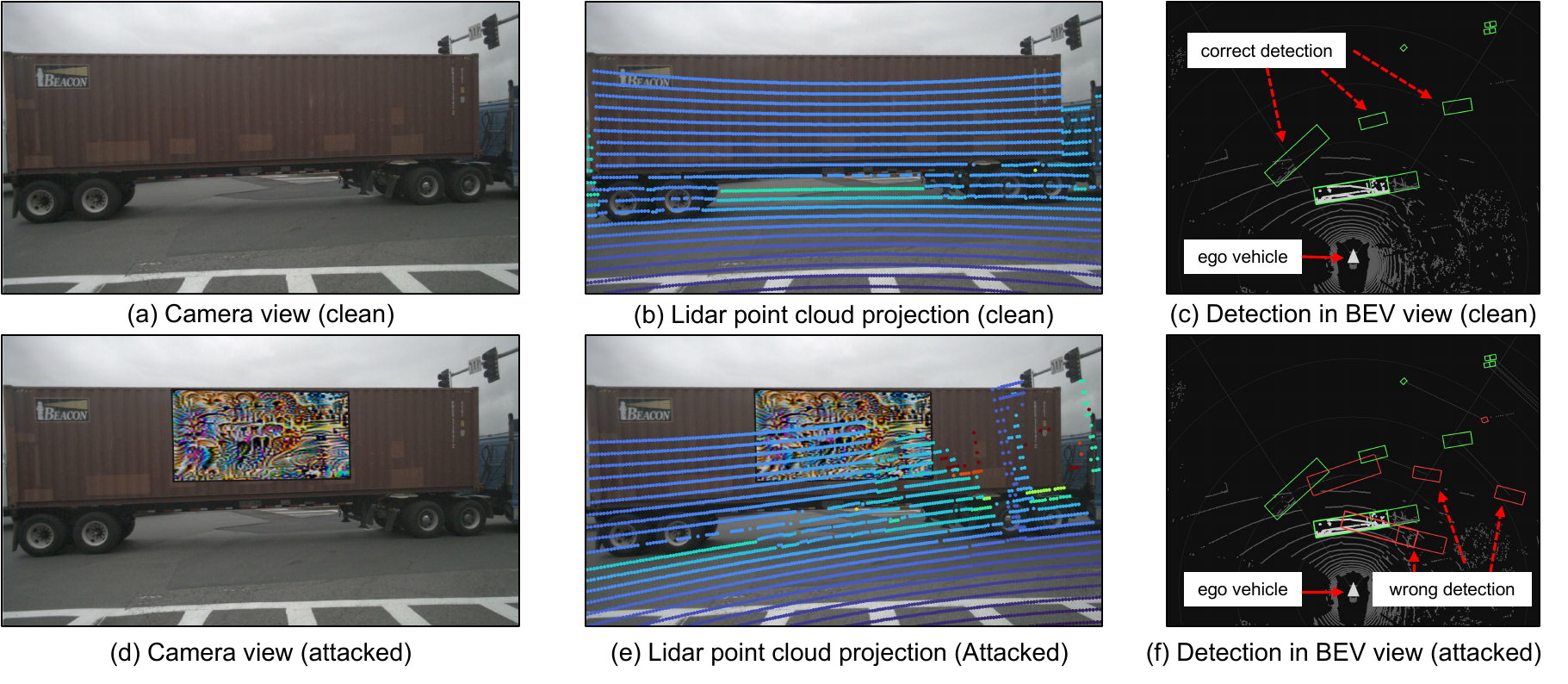}
\caption{An example illustrates the impact of the calibration attack on downstream detection.}
\label{fig:downstream}
\end{figure*}

Camera--LiDAR fusion is a foundation of autonomous-vehicle (AV) perception.
It relies on an extrinsic transform that maps LiDAR points into the camera
frame, governing the cross-modal projection and feature association used by
3D detection, bird's-eye-view (BEV) perception, tracking, and ultimately
planning and control. Although this transform is typically calibrated at the
factory, vibration, thermal expansion, maintenance, and repair can perturb the
relative sensor poses over a vehicle's lifetime~\cite{tahiraj2025calornocal}.
Researchers have therefore developed online calibration
systems~\cite{schneider2017regnet,iyer2018calibnet,lv2021lccnet,
yuan2020rggnet,jing2024surrogatediffusion} that use synchronized images and
point clouds to correct calibration drift while the vehicle is operating.
This capability is also emerging commercially. For example, Spleenlab's
VISIONAIRY Auto Multi Modal Sensor Calibration advertises targetless
camera--LiDAR calibration with continuous or on-demand runtime updates at up
to $5$~Hz~\cite{spleenlab2026onlinecalibration}. Online calibration thus
allows multi-sensor systems to maintain alignment without repeated factory
service.

Making the extrinsic transform updatable at runtime, however, creates a
high-leverage attack surface that has received little attention. Prior AV
security research has studied attacks on camera, LiDAR, and other
modalities~\cite{cao2019adversarial,sun2020lidar,jin2023plalidar}, as well as
multi-sensor fusion models~\cite{hallyburton2022frustum,tu2021msfrobust,
cao2021msfadv,cheng2024fusionnotenough,zhu2024msfattack}, but largely treats
calibration as fixed, accurate, and trusted. Conversely, online calibration
research focuses on estimation accuracy and drift recovery while assuming a
benign physical environment~\cite{schneider2017regnet,iyer2018calibnet,
lv2021lccnet,yuan2020rggnet,jing2024surrogatediffusion}. Missing between these
two lines of work is the adversarial robustness of the calibration process
itself: \emph{can a physical adversarial object trigger recalibration and
cause the system to write an incorrect transform even when the camera and
LiDAR themselves report correct measurements?}

This question matters beyond calibration accuracy. The extrinsic transform is
mutable, safety-critical state shared by the fusion stack. Once corrupted,
it associates otherwise correct LiDAR points with incorrect image regions and
propagates the resulting geometric error to downstream components that consume
the transform. Figure~\ref{fig:downstream} illustrates this effect: the sensor
observations, detector, and 3D boxes are identical in both panels, yet changing
only $T_{\ell \rightarrow c}$ shifts valid projections from vehicles onto the
road, sidewalk, and neighboring objects. Online calibration is therefore not
merely a preprocessing utility, but a robustness boundary between physical
sensing and the rest of the autonomous stack.

Attacking this boundary presents a distinct challenge. The online calibration
pipeline we study is gated: a miscalibration detector first decides
\emph{when} recalibration should run, after which a calibrator determines
\emph{what} extrinsic transform should replace the current
state~\cite{tahiraj2025calornocal,wei2024online}. An attacker must therefore
subvert both decisions using a single physical artifact that is observed
coherently by the camera and LiDAR.

We present \name\ (\emph{\nameLong}), to our knowledge the first physically
realizable attack against the detector--calibrator pipeline of online
camera--LiDAR calibration. \name\ uses a single printable poster placed in the
joint camera--LiDAR field of view. It first spoofs the miscalibration detector
to trigger an unnecessary update and then steers the invoked calibrator toward
an incorrect extrinsic transform. A unified optimization jointly designs the
poster's geometry and texture for both objectives. Its Stage-1 loss is hinged
so that, once the detector fires by a sufficient margin, optimization focuses
on corrupting calibration; for Stage~2, we differentiate through the
calibrator's full iterative trajectory. Universal training across multiple
frames and expectation over transformations (EOT)~\cite{athalye2018eot}
capture scene and placement variation, while a physically constrained renderer
models an opaque printed surface, area-averaged image formation, and LiDAR
returns sampled on the sensor's angular grid. The resulting camera appearance
and LiDAR returns therefore correspond to one physically coherent artifact
rather than independent digital perturbations. This design also exposes a
blind spot in input-consistency defenses that do not independently validate
the extrinsic state written by the calibration
pipeline~\cite{yu2024physense,xu2024physcout,ccs2024visionguard}.

Our evaluation shows that a single adversarial object can simultaneously
compromise the recalibration trigger and the calibration estimator, producing
large persistent calibration errors that translate into downstream safety
consequences. On KITTI, \name\ triggers the miscalibration detector on $89\%$
of held-out frames and induces up to $33.9^\circ$ mean marginal rotation
damage, with $92\%$ of trials exceeding $5^\circ$. The resulting corrupted
extrinsic reduces PointPainting Car 3D AP@R40 from $51.72$ to $0.36$, while
PointPillars, which does not consume the extrinsic, remains essentially
unchanged. Experiments on nuScenes reproduce the calibration-corruption
mechanism under a different sensor suite and coordinate convention. We further
reproduce the attack in closed-loop CARLA, where accepting the corrupted
estimate can cause a collision, and on a physical Clearpath Husky A300 using a
printed adversarial board captured by a real camera--LiDAR rig. Together,
these results demonstrate that online calibration is a practical and
safety-critical attack surface for AVs.

This paper makes the following contributions:
\begin{itemize}[itemsep=0pt, topsep=0pt]
    \item \textbf{A new AV attack surface.} We identify online camera--LiDAR
    calibration as a security-critical, mutable state that has largely remained
    trusted in prior AV security research.

    \item \textbf{A physically realizable calibration attack.} We design a
    two-stage attack in which a single printable artifact both triggers
    recalibration and steers the estimator toward an incorrect extrinsic,
    enabled by joint geometry--texture optimization and physically constrained
    rendering.

    \item \textbf{A multi-level system evaluation.} We demonstrate calibration
    corruption across KITTI and nuScenes, quantify its impact on camera--LiDAR
    fusion, reproduce it on a physical Husky robot, and show closed-loop safety
    consequences in CARLA.
\end{itemize}

\section{Background and Related Work}
\label{sec:background}


\vspace{1mm}
\noindent\textbf{Camera--LiDAR fusion and extrinsic calibration.}
Modern autonomous vehicles (AVs) combine cameras and LiDAR to exploit their
complementary sensing capabilities: cameras provide dense appearance and semantic
information, while LiDAR provides accurate 3D geometry.
Production AV platforms such as Apollo and Autoware~\cite{baiduapollo,autoware}
use these modalities for tasks including 3D object detection, BEV perception,
tracking, and planning.
Cross-modal fusion relies on the camera--LiDAR extrinsic
$T_{\ell\rightarrow c}$, a rigid transformation that maps points from the LiDAR
coordinate frame into the camera frame.
Once $T_{\ell\rightarrow c}$ is given, operations such as projecting LiDAR
points into the image, associating image features with 3D points, and constructing
cross-modal BEV representations follow deterministically.
Consequently, an erroneous extrinsic can systematically corrupt fusion even when
both sensors themselves provide correct measurements.

Camera--LiDAR extrinsics are typically initialized offline using calibration
targets or scene correspondences.
However, vibration, thermal variation, maintenance, and small mechanical
displacements can change the relative sensor pose during deployment
~\cite{tahiraj2025calornocal,wei2024online}.
Online calibration therefore re-estimates the extrinsic during operation,
avoiding repeated manual or factory calibration.
Learning-based approaches such as RegNet~\cite{schneider2017regnet},
CalibNet~\cite{iyer2018calibnet}, LCCNet~\cite{lv2021lccnet},
RGGNet~\cite{yuan2020rggnet}, and surrogate diffusion
~\cite{jing2024surrogatediffusion} infer the relative camera--LiDAR pose from
synchronized images and point clouds, often through iterative refinement of an
initial estimate.
Unlike a transient perception output, the resulting extrinsic becomes
\emph{persistent system state}: once accepted, it can be reused by many subsequent
fusion operations until another calibration is performed.

\vspace{1mm}
\noindent\textbf{Miscalibration detection and runtime update.}
Online calibration need not run continuously.
Existing systems can first determine whether the currently deployed extrinsic
has drifted and invoke recalibration only when necessary
~\cite{tahiraj2025calornocal,wei2024online}.
Such detectors estimate cross-modal inconsistency using cues such as projected
LiDAR geometry, image structure, or global relative displacement.
The resulting pipeline therefore contains two security-critical decisions:
\emph{when} an extrinsic should be updated and \emph{what} new transformation
should replace it.
This gating structure is important because falsely triggering calibration alone
does not corrupt the vehicle, while corrupting a calibrator is ineffective if
the calibrator is never invoked.
More fundamentally, both stages infer correctness from the same physical
camera--LiDAR observations that an adversarial object can influence.
This creates a previously underexplored attack surface at the boundary between
physical sensing and sensor fusion.

\vspace{1mm}
\noindent\textbf{Physical attacks on individual sensors and perception tasks.}
Prior physical AV attacks primarily manipulate sensor observations or the
perception models consuming them.
LiDAR attacks inject, remove, or alter returns through optical, electromagnetic,
or physical mechanisms
~\cite{petit2015remote,shin2017illusion,cao2019adversarial,sun2020lidar,
takami2025mvslidar,jin2025phantomlidar,jin2023emilidar,jin2024newgenlidar,
jin2023plalidar,cao2023youcantseeme,kobayashi2025shadow}.
Camera-side attacks use printed patterns, projected light, infrared illumination,
or other physical stimuli to manipulate traffic-sign recognition, object
detection, monocular depth, visual SLAM, lane detection, and steering
~\cite{zhao2019seeing,lovisotto2021slap,wang2025controlloc,ndss2025tsr,
anon2022signpatch,kohli2024invisiblereflections,yan2022rollingcolors,
guo2024ghoststripe,cheng2022mdepatch,zheng2024_3d2fool,zheng2024pijack,
chen2024aor,sato2021dirty,jing2021toogood,zhou2020deepbillboard}.
Physical-object attacks further optimize object geometry or appearance so that
real objects induce adversarial LiDAR or camera observations
~\cite{tu2020physically,zhu2024ae,zhu2021can}.
These attacks establish the feasibility of modifying the physical environment,
but their objective is to directly corrupt sensor data or a downstream
recognition task.

\vspace{1mm}
\noindent\textbf{Attacks on multi-sensor fusion.}
A second line of work attacks the point at which multiple modalities are fused.
Prior work has constructed 3D-printable objects that fool camera--LiDAR fusion
detectors~\cite{cao2021msfadv}, injected LiDAR points into camera-defined
frustums~\cite{hallyburton2022frustum}, designed modality-consistent physical
attacks~\cite{tu2021msfrobust}, shown that a single modality can compromise
fusion~\cite{cheng2024fusionnotenough}, and extended attacks to additional
modalities and collaborative perception
~\cite{zhu2024msfattack,zhang2024advcollab,wang2025sombra,lou2025onlinemap}.
Related localization attacks manipulate GPS or SLAM observations
~\cite{shen2020drift,zhang2025ghostnav,wang2021invisible}.
Despite differing attack mechanisms, these works share an important assumption:
the geometric transformations connecting sensors are fixed and trustworthy.
The attacker manipulates what the sensors observe or what the fusion model
computes, but not the extrinsic that determines how otherwise correct
measurements are geometrically associated.

\name\ targets a different layer.
It adopts the physical-object threat model of prior work, but uses the object
to manipulate the module that \emph{produces} $T_{\ell\rightarrow c}$.
The resulting corruption is therefore upstream of fusion rather than an attack
on a particular fusion architecture.
Once the incorrect transformation is accepted, downstream components consume
misregistered yet otherwise legitimate sensor measurements without requiring
continued modification of their inputs.

\section{Attack Model}
\label{sec:attack-model}

\begin{figure}[t]
\centering
\includegraphics[width=\columnwidth]{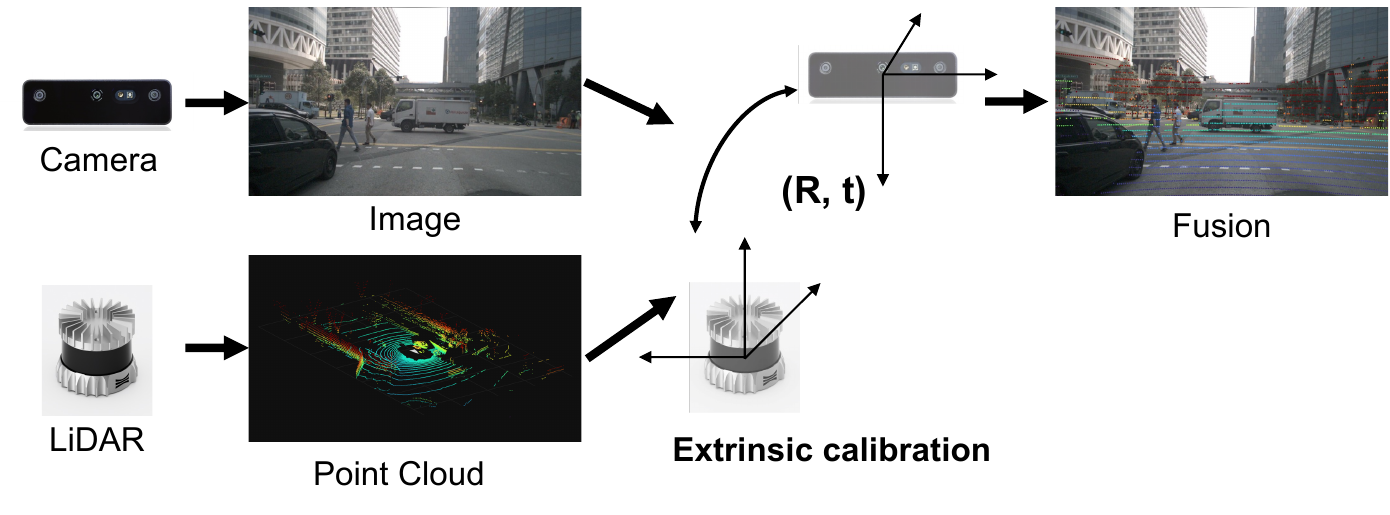}
\caption{Extrinsic calibration (rotation $R$ and translation $t$) between camera and LiDAR for sensor fusion.}
\label{fig:RT-example}
\end{figure}

\name\ targets a layer of the perception stack left untouched
by prior LiDAR-spoofing and 3D-adversarial-object attacks: 
the camera--LiDAR \emph{extrinsic calibration module} that 
produces a LiDAR-to-camera transform on which every downstream 
fusion operation relies.
Once adopted, this corrupted extrinsic propagates into every 
subsequent BEV-occupancy, 3D-detection, and tracking step, yielding
an infrastructure-level cross-modal misalignment that is invisible 
to sensor- and detector-level defenses while each sensor 
continues to report honest data.

\subsection{Attack Target System}
\label{sec:atk-system}


We consider an AV equipped with a forward-facing RGB camera and
a 3D LiDAR rigidly mounted on the vehicle's rooftop. 
Fig.~\ref{fig:RT-example} shows the extrinsic calibration 
between camera and LiDAR for sensor fusion.
The perception stack maintains a current extrinsic estimate
$T_{\ell \rightarrow c}^{(0)} \in SE(3)$ at every frame:
\[
T_{\ell \rightarrow c}^{(0)} =
\begin{bmatrix} R & t \\ 0 & 1 \end{bmatrix},
\]
where $R \in SO(3)$ and $t \in \mathbb{R}^3$.
Two learned modules govern how $T_{\ell \rightarrow c}^{(0)}$
is updated at deployment time:
\begin{enumerate}[leftmargin=*, itemsep=0pt, topsep=2pt, parsep=0pt]
  \item A self-supervised \emph{miscalibration detector} 
    $\mathcal{D}(I, P; T_{\ell \rightarrow c}^{(0)}) \in [0,1]$ 
    that assesses how badly the current extrinsic aligns the
    projected LiDAR depth map with the camera image.
    Online calibration is invoked only when $\mathcal{D}$ exceeds a
    drift threshold $\tau_{\mathrm{det}}$~\cite{tahiraj2025calornocal, wei2024online}.
  \item A \emph{calibration network} $f_\theta$ that, when invoked, 
    refines $T_{\ell \rightarrow c}^{(0)}$ via an iterative update:
    \begin{equation}
    \label{eq:iter-refine}
    T_{\ell \rightarrow c}^{(k+1)} = \exp\!\big(\Delta \xi^{(k)}\big)\, 
    T_{\ell \rightarrow c}^{(k)},
    \qquad k = 0, \ldots, K-1,
    \end{equation}
    and outputs a final estimate $\hat{T}_{\ell \rightarrow c} = T_{\ell 
    \rightarrow c}^{(K)}$ that downstream BEV fusion, detection, 
    and tracking consume~\cite{schneider2017regnet,iyer2018calibnet,
    lv2021lccnet,yuan2020rggnet,jing2024surrogatediffusion}.
\end{enumerate}
The gating structure $\{\mathcal{D}, f_\theta\}$ together 
determines \emph{when} an extrinsic update happens and 
\emph{to what value} it converges.
Both must be subverted for an attacker to corrupt the AV's 
extrinsic parameters.

\subsection{Attacker's Capabilities and Goals}
\label{sec:atk-goal}

The attacker aims to compromise the real-time camera--LiDAR 
auto-calibration pipeline using a single physically realizable 
cross-modal object. The object is designed to first induce 
sufficient cross-modal inconsistency to trigger automatic 
recalibration and then create misleading visual--geometric 
correspondences that steer the calibrator toward an incorrect 
extrinsic estimate. The resulting calibration error persists 
in subsequent sensor-fusion operations until the system 
calibrates again or restores a trusted extrinsic.

The attacker can design the physical object and choose 
its deployment location, providing partial control over the
local attack scenario. We model the adversarial object generally 
as any physical surface or structure observable by both camera 
and LiDAR, with realizable examples including a planar adversarial 
object on the roadside or a poster mounted on the rear of a truck. 
The attacker cannot predict the victim's exact motion, viewpoint, 
or sensing conditions, so the object should remain effective under
such runtime variations at the chosen site. Generalization across 
arbitrary road segments or deployment locations is unnecessary 
because the attacker selects the deployment site. 
The object must also satisfy practical constraints on size, 
placement, LiDAR returns, and physical constructibility.

\subsection{Attacker's Knowledge}
\label{sec:atk-knowledge}

We assume a \emph{matched-surrogate gray-box} adversary with 
the following knowledge:
\begin{itemize}[itemsep=0pt, topsep=2pt, parsep=0pt]
    \item The architectural family of the deployed miscalibration 
      detector $\mathcal{D}$ and calibration network $f_\theta$, 
      and their iterative-refinement scheme (Eq.~\ref{eq:iter-refine}).
    \item The sensor configuration, including camera intrinsics, 
      approximate mounting locations of camera and LiDAR, and their 
      overlapping fields of view.
    \item The operating data distribution (road geometry, typical 
      ranges, motion patterns), and the ability to train a surrogate 
      calibration network $f_{\theta_s}$ on similar data.
\end{itemize}
The attacker does \emph{not} require the exact parameters of 
the deployed $\mathcal{D}$ or $f_\theta$ at runtime, nor any access
to their internal states.
At optimization time, gradients flow through the attacker-trained 
surrogate $f_{\theta_s}$; at deployment time, the attacker only 
needs to place the optimized artifact in the environment.

\subsection{Trusted Components}
\label{sec:atk-trusted}

The following components of the AV stack are outside our threat
model and assumed uncompromised:
\begin{itemize}[itemsep=0pt, topsep=2pt, parsep=0pt]
    \item The vehicle's control stack (planning and control) and 
      in-vehicle networks (e.g., CAN, automotive Ethernet).
    \item The camera and LiDAR hardware, including their 
      synchronization, firmware, and noise-rejection logic.
    \item The implementation of the calibration network $f_\theta$ 
      and the miscalibration detector $\mathcal{D}$ (the attacker
      cannot modify their code, weights, or internal states at runtime).
    \item The operating system and any signed software on the vehicle.
\end{itemize}
The only exposed attack surface is the sensor input space --- the 
physical environment observed by the sensors at deployment time,
and, for the digital baseline only, the raw data buffer between 
the sensors and $\mathcal{D}/f_\theta$.
No tampering with the sensors themselves, the in-vehicle networks,
or the calibration code is required or permitted by the threat model.

\section{Attack Design}
\subsection{Overview}
\label{sec:attack-overview}

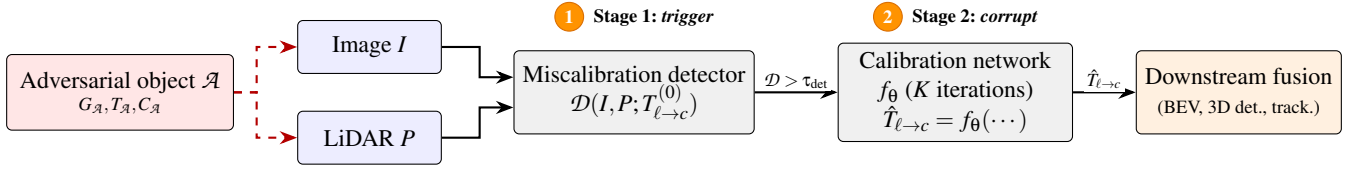
\begin{figure*}[t]
\centering
\begin{tikzpicture}[
>=Stealth,
font=\small,
sensor/.style={draw, rounded corners=2pt, align=center,
minimum height=7mm, minimum width=19mm, fill=blue!7},
avmod/.style={draw, rounded corners=2pt, align=center,
minimum height=11mm, minimum width=31mm, fill=gray!12, font=\small},
attack/.style={draw, rounded corners=2pt, align=center,
minimum height=10mm, minimum width=30mm, fill=red!10, font=\small},
downstream/.style={draw, rounded corners=2pt, align=center,
minimum height=11mm, minimum width=27mm, fill=orange!12, font=\small},
flow/.style={-Stealth, thick},
adv/.style={-Stealth, thick, red!70!black, dashed},
edgelbl/.style={font=\scriptsize, fill=white, inner sep=1pt},
stagelbl/.style={font=\scriptsize\bfseries, align=center},
badge/.style={circle, draw=orange!85!black, fill=orange!85!yellow,
text=white, font=\scriptsize\bfseries,
inner sep=0pt, minimum size=3.8mm}
]

\node[attack] (art) at (-2.8,0)
{Adversarial object $\mathcal{A}$\\[-1mm]
\scriptsize $G_{\mathcal{A}}, T_{\mathcal{A}}, C_{\mathcal{A}}$};

\node[sensor] (img) at (0.5, 0.6) {Image $I$};
\node[sensor] (pcd) at (0.5,-0.6) {LiDAR $P$};

\node[avmod] (det) at (4,0)
{Miscalibration detector\\
$\mathcal{D}(I,P;T_{\ell\to c}^{(0)})$};

\node[avmod] (cal) at (8.25,0)
{Calibration network\\
$f_\theta$ ($K$ iterations)\\
$\hat{T}_{\ell\to c}=f_\theta(\cdots)$};

\node[downstream] (down) at (12,0)
{Downstream fusion\\
\scriptsize (BEV, 3D det., track.)};

\node[stagelbl] (s1lbl) at (4,1.0)
{\tikz[baseline=-0.6ex]\node[badge]{1};\; Stage 1: \emph{trigger}};

\node[stagelbl] (s2lbl) at (8.25,1.0)
{\tikz[baseline=-0.6ex]\node[badge]{2};\; Stage 2: \emph{corrupt}};

\draw[flow] (img.east) -- ([xshift=5mm]img.east) |- ([yshift=2mm]det.west);
\draw[flow] (pcd.east) -- ([xshift=5mm]pcd.east) |- ([yshift=-2mm]det.west);
\draw[flow] (det.east) --
node[edgelbl, above]{$\mathcal{D}>\tau_{\mathrm{det}}$} (cal.west);
\draw[flow] (cal.east) --
node[edgelbl, above]{$\hat{T}_{\ell\to c}$} (down.west);

\draw[adv] (art.east) -- ([xshift=3mm]art.east) |- (img.west);
\draw[adv] (art.east) -- ([xshift=3mm]art.east) |- (pcd.west);

\end{tikzpicture}
\caption{End-to-end pipeline of \name. A single physical object
$\mathcal{A}$ affects both camera and LiDAR observations. 
\textbf{Stage~1} pushes the miscalibration score above 
$\tau_{\mathrm{det}}$ to trigger online recalibration. 
\textbf{Stage~2} then steers the invoked calibrator toward an
incorrect extrinsic $\hat{T}_{\ell\to c}$, which is subsequently 
consumed by downstream fusion.
}
\label{fig:pipeline}
\end{figure*}

\name\ is a two-stage physical attack against an AV's online
camera--LiDAR calibration pipeline. 
As shown in Fig.~\ref{fig:pipeline}, a single adversarial 
object $\mathcal{A}$ in the joint camera--LiDAR field of view 
first \emph{triggers} unnecessary recalibration and then 
\emph{corrupts} the resulting extrinsic estimate.
Let $I^{\mathcal{A}}$ and $P^{\mathcal{A}}$ denote the camera 
image and LiDAR point cloud observed in the presence of $\mathcal{A}$.

In Stage~1, the object must induce sufficient apparent cross-modal
inconsistency to trigger the miscalibration detector:
\begin{equation}
\label{eq:goal-trigger}
\mathcal{D}\!\left(
I^{\mathcal{A}},
P^{\mathcal{A}};
T_{\ell \rightarrow c}^{(0)}
\right)
>
\tau_{\mathrm{det}},
\end{equation}
where $T_{\ell \rightarrow c}^{(0)}$ is the currently deployed 
extrinsic and $\tau_{\mathrm{det}}$ is the detector threshold. 
Without satisfying this condition, the calibration module 
is never invoked.

In Stage~2, the same object must steer the invoked calibrator toward 
an incorrect transform $\hat{T}_{\ell\rightarrow c}^{\mathcal{A}}$.
We quantify its deviation from the true extrinsic 
$T^\star_{\ell\rightarrow c}$ using
\begin{equation}
\label{eq:calib-error}
\begin{aligned}
\mathcal{E}\!\left(
\hat{T}_{\ell \rightarrow c},
T^\star_{\ell \rightarrow c}
\right)
&=
\left\|\log(\Delta R)\right\|_2
+
\lambda_t \left\|\Delta t\right\|_2, \\
\Delta T
&=
\left(T^\star_{\ell \rightarrow c}\right)^{-1}
\hat{T}_{\ell \rightarrow c}
=
\begin{bmatrix}
\Delta R & \Delta t \\
0 & 1
\end{bmatrix},
\end{aligned}
\end{equation}
where $\Delta R$ and $\Delta t$ denote the rotational and translational
calibration errors, respectively, and $\lambda_t>0$ balances the two terms.

The key challenge is that the two stages cannot rely on independent
digital perturbations. A \emph{single physical object} must 
simultaneously provide the evidence that triggers recalibration and 
the misleading correspondences that corrupt its output. 
\name, therefore, jointly optimizes one physically
realizable appearance against both objectives.

\subsection{Cross-Modal Rendering of the Adversarial 
   Object for Optimization}
\label{sec:artifact-design}


We model the adversarial object as
\begin{equation}
\label{eq:artifact}
\mathcal{A}
=
\left(
G_{\mathcal{A}},
T_{\mathcal{A}},
C_{\mathcal{A}}
\right),
\end{equation}
where $G_{\mathcal{A}}$ denotes its physical geometry,
$T_{\mathcal{A}}$ its deployment pose, and $C_{\mathcal{A}}$ the 
appearance applied to a flat or approximately flat surface. 
This abstraction is not tied to a particular object type. 
For example, the surface may be a roadside adversarial object,
a poster mounted on the rear of a truck, or part of another
physical structure visible to both camera and LiDAR.

For a chosen deployment, $G_{\mathcal{A}}$ and the nominal
$T_{\mathcal{A}}$ are fixed, while $C_{\mathcal{A}}$ is optimized. 
This matches the attack model: the attacker can construct and 
place the object before deployment, but cannot electronically 
manipulate either sensor as the victim approaches.

\begin{figure}[!ht]
\centering
\includegraphics[width=.8\columnwidth]{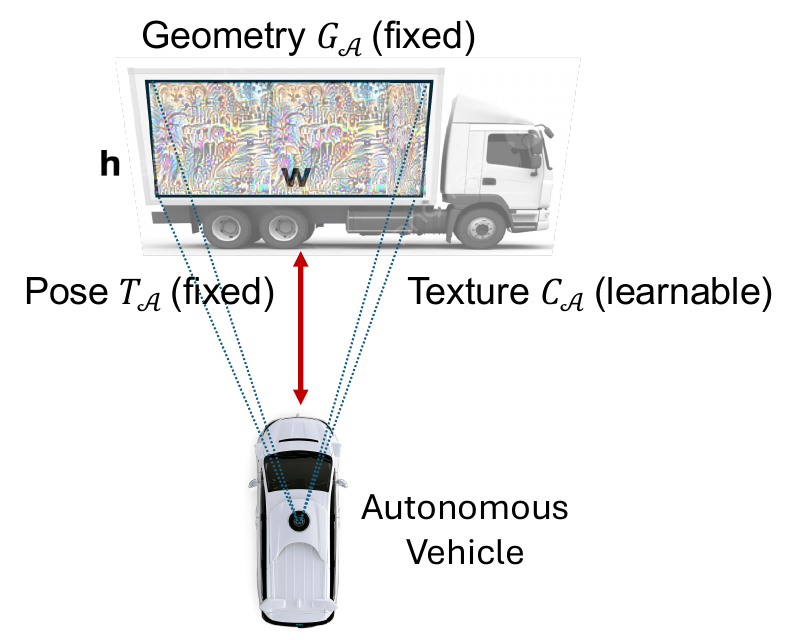}
\caption{\textbf{Adversarial-object parameterization and one 
planar realization.} We represent $\mathcal{A}$ using
(a)~physical geometry $G_{\mathcal{A}}$,
(b)~deployment pose $T_{\mathcal{A}}$, and
(c)~surface appearance $C_{\mathcal{A}}$.
The illustrated planar panel is one realization of the general 
object model; the same formulation applies to other physical 
objects exposing a printable surface jointly observable by 
camera and LiDAR. For a chosen deployment,
$G_{\mathcal{A}}$ and the nominal $T_{\mathcal{A}}$ are fixed, 
while $C_{\mathcal{A}}$ carries both attack stages and is optimized.}
\label{fig:artifact-param}
\end{figure}

A central requirement is that the attacked camera and LiDAR observations
correspond to the \emph{same} physical object. We therefore define a
cross-modal observation operator
\begin{equation}
\label{eq:render}
(I^{\mathcal{A}},P^{\mathcal{A}})
=
\mathcal{R}\!\left(
I,P;
G_{\mathcal{A}},
T_{\mathcal{A}},
C_{\mathcal{A}}
\right).
\end{equation}

On the camera side, the visible surface of $\mathcal{A}$ is projected 
using the true extrinsic $T^\star_{\ell\rightarrow c}$ and rendered as 
an \emph{opaque} surface. 
The pixels covered by the object are therefore replaced by its 
appearance rather than blended with the background. 
This prevents the optimizer from exploiting transparency or additive 
intensity patterns that a real printed object cannot produce.

On the LiDAR side, the same geometry produces returns according to the
sensor's angular sampling pattern. The camera texture and LiDAR returns 
are therefore generated from a shared physical surface rather than 
optimized as independent perturbations. Camera appearance resolution
and LiDAR sampling density are also represented separately, avoiding 
an artificial one-to-one mapping between image texels and LiDAR returns.

The observation operator is differentiable with respect to
$C_{\mathcal{A}}$. Gradients from both attack stages can therefore 
propagate through sensor formation to the same physical appearance.

\subsection{Stage 1: Spoof the Miscalibration Detector}
\label{sec:stage1}

The first stage targets the detector that determines whether online
recalibration should run or not. We abstract it as a learned 
consistency function $\mathcal{D}(I,P;T_{\ell\to c}^{(0)})$ that
evaluates whether the camera and LiDAR observations agree under
the current extrinsic. Existing online calibration systems commonly
detect drift from discrepancies between image features and projected
LiDAR geometry~\cite{tahiraj2025calornocal,wei2024online}.

The attack minimizes the hinge loss
\begin{equation}
\label{eq:trigger}
\mathcal{L}_{\mathrm{trig}}(\mathcal{A})
=
\max\!\Big(
0,\;
\tau_{\mathrm{det}} + m -
\mathcal{D}\!\left(
I^{\mathcal{A}},
P^{\mathcal{A}};
T_{\ell\to c}^{(0)}
\right)
\Big),
\end{equation}
where $m>0$ is a trigger margin. The loss pushes the detector above
its threshold and becomes zero once it exceeds the threshold by $m$.

The hinge is important because the detector is only a gate.
Once recalibration is reliably triggered, further increasing its
score provides no additional attack benefit. 
Clamping this objective allows the remaining optimization capacity
to focus on corrupting the resulting calibration estimate.

\subsection{Stage 2: Corrupt the Extrinsic Estimate}
\label{sec:stage2}

Once Stage~1 triggers recalibration, the same physical object steers 
the calibrator away from the true extrinsic. 
Following the attac model in Section~\ref{sec:attack-model},
the attacker optimizes through a differentiable surrogate 
calibrator $f_{\theta_s}$ and evaluates transfer to
the deployed calibrator separately.

For an iterative calibrator, the estimate is refined as
\begin{equation}
\label{eq:iter-refine}
T^{(k+1)}
=
\exp\!\left(\Delta\xi^{(k)}\right)T^{(k)},
\qquad
k=0,\ldots,K-1.
\end{equation}
The artifact therefore affects not only one pose prediction, 
but the entire refinement trajectory.
Optimizing against only the first update may produce a temporary 
error that later iterations correct. We instead unroll all $K$
iterations and optimize the final estimate:
\begin{equation}
\label{eq:corrupt}
\mathcal{L}_{\mathrm{corr}}(\mathcal{A})
=
-\mathcal{E}\!\left(
\hat{T}_{\ell\to c}^{\mathcal{A}}(f_{\theta_s}),
T^\star_{\ell\to c}
\right).
\end{equation}

Gradients from the final calibration error propagate through the
complete refinement trajectory, then through the attacked camera 
observation and ultimately to $C_{\mathcal{A}}$. The optimization 
therefore targets the calibrator's converged output rather than an 
intermediate prediction.

\subsection{Joint and Robust Optimization}
\label{sec:joint-opt}

The trigger and corruption objectives share the same physical appearance.
Optimizing them independently does not guarantee that the resulting 
objects compose: an object that strongly corrupts calibration is 
ineffective if it does not first trigger recalibration, while an object 
optimized only to fire the detector need not bias the calibration estimate.

We therefore optimize both stages jointly:
\begin{equation}
\label{eq:joint}
\min_{C_{\mathcal{A}}}
\quad
\mathcal{L}_{\mathrm{corr}}(\mathcal{A})
+
\lambda_{\mathrm{trig}}
\mathcal{L}_{\mathrm{trig}}(\mathcal{A}),
\end{equation}
with $G_{\mathcal{A}}$ and the nominal $T_{\mathcal{A}}$ fixed for the
chosen deployment. The trigger loss ensures that the calibration 
routine is invoked, while the corruption loss maximizes the error 
in the state it writes.

\begin{figure}[t]
\centering
\begin{tikzpicture}[
    scale=0.9, transform shape,
    >=Stealth, font=\small,
    art/.style={draw, rounded corners=2pt, fill=red!10, align=center,
                 minimum width=20mm, minimum height=8mm, font=\small},
    scene/.style={draw, rounded corners=2pt, fill=blue!7, align=center,
                  minimum width=22mm, minimum height=8mm, font=\scriptsize},
    block/.style={draw, rounded corners=2pt, fill=gray!12, align=center,
                  minimum width=25mm, minimum height=9mm, font=\scriptsize},
    loss/.style={draw, rounded corners=2pt, fill=orange!15, align=center,
                 minimum width=27mm, minimum height=8mm, font=\scriptsize},
    total/.style={draw, rounded corners=2pt, fill=green!12, align=center,
                  minimum width=44mm, minimum height=9mm, font=\small},
    flow/.style={-Stealth, thick},
    grad/.style={-Stealth, thick, blue!70!black, dashed},
]

\node[art] (art) at (0, 4)
{Object $\mathcal{A}$\\
\scriptsize $G_\mathcal{A},T_\mathcal{A},C_\mathcal{A}$};

\node[scene] (sceneL) at (-2.6, 2.7)
{$I^\mathcal{A},P^\mathcal{A}$};
\node[scene] (sceneR) at ( 2.6, 2.7)
{$I^\mathcal{A},P^\mathcal{A}$};

\node[block] (det) at (-2.6, 1.5)
{Detector $\mathcal{D}$\\one forward pass};
\node[block] (cal) at ( 2.6, 1.5)
{Calibrator $f_{\theta_s}$\\$K$ unrolled iterations};

\node[loss] (ltrig) at (-2.6, 0.2)
{$\mathcal{L}_{\mathrm{trig}}
=\max(0,\tau{+}m{-}\mathcal{D})$};
\node[loss] (lcorr) at ( 2.6, 0.2)
{$\mathcal{L}_{\mathrm{corr}}
=-\mathcal{E}(\hat{T}_{\ell\to c}^{\mathcal{A}},T^\star)$};

\node[total] (tot) at (0, -1.1)
{$\mathcal{L}
=\mathcal{L}_{\mathrm{corr}}
+\lambda_{\mathrm{trig}}\mathcal{L}_{\mathrm{trig}}$};

\draw[flow] (art) -- (sceneL);
\draw[flow] (art) -- (sceneR);
\draw[flow] (sceneL) -- (det);
\draw[flow] (sceneR) -- (cal);
\draw[flow] (det)   -- (ltrig);
\draw[flow] (cal)   -- (lcorr);
\draw[flow] (ltrig) -- (tot);
\draw[flow] (lcorr) -- (tot);

\draw[grad] (tot.west) -| (-5.0,4.0) -- (art.west);
\node[font=\scriptsize, blue!70!black, anchor=west]
at (-4.85,2.2)
{$\partial\mathcal{L}/\partial C_\mathcal{A}$};

\node[font=\scriptsize\bfseries, orange!85!black]
at ([yshift=5mm]sceneL.north)
{Stage 1: \emph{trigger}};
\node[font=\scriptsize\bfseries, orange!85!black]
at ([yshift=5mm]sceneR.north)
{Stage 2: \emph{corrupt}};

\end{tikzpicture}
\caption{Joint two-stage optimization. The same attacked observations 
feed the miscalibration detector and the unrolled calibration surrogate, 
producing the trigger and corruption losses, respectively.
Both losses back-propagate to the same physical appearance 
$C_{\mathcal A}$. The hinged Stage~1 loss
becomes inactive once recalibration is triggered with sufficient margin,
allowing optimization to focus on Stage~2.
}
\label{fig:two-stage-loss}
\end{figure}
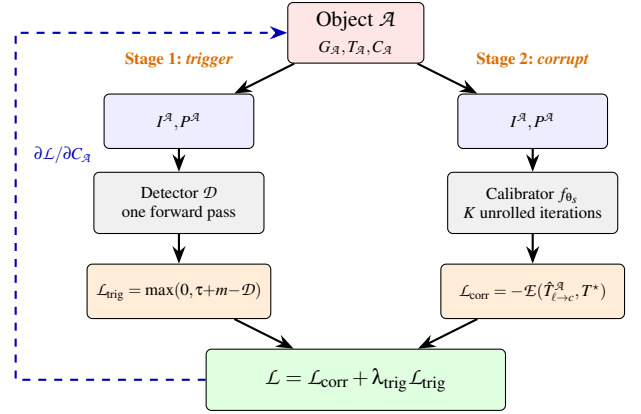

A physically deployed object must also tolerate uncertainty in how 
the victim encounters it. Although the attacker selects the object 
and its deployment site, it cannot know the victim's exact trajectory, 
distance, or viewpoint at runtime. We therefore optimize for 
\emph{deployment robustness} in two ways.

First, instead of optimizing against a single observation, we optimize 
one appearance over a set of frames $\mathcal{F}_{\mathrm{train}}$ 
representing different vehicle positions and viewpoints near the 
deployment location.
This produces a universal object for the chosen scenario rather 
than a frame-specific perturbation.

Second, we apply expectation over transformations
(EOT)~\cite{athalye2018eot} to model additional uncertainty in the
object--vehicle relationship. 
For each sampled frame, we perturb the nominal
deployment pose by $\delta T\sim\mathcal{T}$ and reconstruct 
the corresponding cross-modal observations:
\begin{equation}
\label{eq:eot-render}
(I^{\mathcal{A}},P^{\mathcal{A}})
=
\mathcal{R}\!\left(
I,P;
G_{\mathcal{A}},
T_{\mathcal{A}}\delta T,
C_{\mathcal{A}}
\right).
\end{equation}

The complete optimization becomes
\begin{equation}
\label{eq:eot-joint}
\min_{C_{\mathcal{A}}}
\;
\mathbb{E}_{(I,P)\sim\mathcal{F}_{\mathrm{train}},
\,\delta T\sim\mathcal{T}}
\left[
\mathcal{L}_{\mathrm{corr}}(\mathcal{A})
+
\lambda_{\mathrm{trig}}
\mathcal{L}_{\mathrm{trig}}(\mathcal{A})
\right].
\end{equation}

For each optimization step, we sample driving observations and pose
perturbations, generate physically consistent camera--LiDAR observations,
evaluate both stages, and back-propagate the averaged loss through the
detector, the full iterative calibrator, and the cross-modal observation
operator to update $C_{\mathcal{A}}$.

This formulation intentionally targets robustness within a chosen 
deployment scenario rather than universality across arbitrary locations. 
The attacker controls where the object is deployed but not the victim 
vehicle's exact runtime motion; accordingly, \name\ optimizes one 
physical object to tolerate the expected viewpoint and sensing 
variations around that deployment.

\vspace{-2mm}
\section{Implementation}
\label{sec:implementation}

This section describes how the two-stage attack is
realized as a concrete physical artifact and training pipeline.
All experiments in Section~\ref{sec:evaluation} use the 
configuration described below.

\subsection{Surrogate Calibration Network}
\label{sec:impl-surrogate}

The attacker's surrogate $f_{\theta_s}$ is an iterative
refinement network with $K = 10$ refinement steps following 
Eq.~(\ref{eq:iter-refine}).
We use a publicly available learned camera--LiDAR calibrator, 
trained on the KITTI odometry split~\cite{jing2024surrogatediffusion}, 
as the white-box matched surrogate for our main experiments.
Specifically, it is the surrogate-diffusion framework instantiated with
the CalibNet~\cite{iyer2018calibnet} backbone. CalibNet \emph{is} the 
white-box surrogate rather than a separate transfer target: 
the framework and its backbone are one (not two) network.
The surrogate is treated as a black box during forward inference 
but its $K$-step unrolled refinement is fully differentiable 
during attack optimization, so gradients flow from the final-pose 
error all the way back to the artifact's color buffer.
A print-robustness variant of the artifact additionally places 
LCCNet~\cite{lv2021lccnet} inside the objective, making the attack 
white-box on that calibrator as well. RGGNet~\cite{yuan2020rggnet} is the 
only architecture that is genuinely held out --- it never appears in any 
objective --- and it is the transfer target we report in 
Section~\ref{sec:transferability}.

\subsection{Artifact Representation}
\label{sec:impl-artifact}

The physical artifact is a planar rectangular adversarial object of 
$W = 3.6$\,m $\times H = 1.8$\,m, the scale of a roadside advertising panel.
We represent the artifact in the LiDAR frame as a point set $P_{\mathcal{A}} = 
\{p_i\}_{i=1}^{N_{\mathcal{A}}}$ with $N_{\mathcal{A}} = 4{,}100$ points 
laid out on a $164 \times 25$ angular grid over its surface. This grid is fixed 
by measurement rather than chosen: across $88{,}309$ inter-return gaps measured 
on KITTI Velodyne data, a vertical surface at $\sim 6$\,m returns points 
spaced $2.64$\,px horizontally and $8.78$\,px vertically in the camera image,
and $164 \times 25$ is the grid that reproduces that spacing over a
$3.6 \times 1.8$\,m board. A uniform lattice chosen for convenience is 
much finer --- an $89 \times 90$ grid gives $2.40$\,px vertically, roughly
$3.7\times$ more than an HDL-64E can resolve at that range --- and 
substituting a datasheet-average beam spacing is not sufficient either, 
since it gives $5.27$\,px.
The printed texture is a separate $89 \times 90$ texel RGB field $C_{\mathcal{A}}$,
each texel $c \in [0,1]^3$ covering $40.4 \times 20.0$\,mm at $1{:}1$ scale, 
emitted as a $3600 \times 1800$\,px print file at $1$\,mm/px.
The depth side samples this field at the $4{,}100$ return positions and 
the image side rasterizes the board from it, so printed resolution and
LiDAR return density are no longer tied to a single grid.
The adversarial object's pose $T_{\mathcal{A}} \in SE(3)$ is fixed at a chosen 
ground-supported placement ahead of the vehicle (board plane perpendicular 
to the sensor forward axis, centre at LiDAR height, $6.0$\,m from the sensor 
and on-axis) and is not optimized; only the texture $C_{\mathcal{A}}$ is learnable.
This choice reflects the physical reality that a deployed adversarial 
poster has fixed geometry and pose once printed and mounted.

\vspace{-2mm}
\subsection{Sensor Compositing}
\label{sec:impl-sensor}

At each forward step of the attack optimization, we synthesize the AV's image
and point cloud as the AV would observe them with $\mathcal{A}$ present 
in the scene. The artifact point set $P_{\mathcal{A}}$ is added to the 
scene LiDAR point cloud $P$ in the LiDAR frame.
For the camera image, the board is projected through the \emph{true} extrinsic 
$T^\star_{\ell \rightarrow c}$ into the image plane and rendered differentiably 
as an \emph{opaque} planar occluder: every covered pixel takes an area-averaged 
sample of the texture, and the underlying scene pixel is replaced rather than
blended with. Opacity is not a detail. A Gaussian point splat with the compositing 
rule $\big(\mathrm{base} + \sum_i \alpha_i w_i c_i\big) \big/ \sum_i \alpha_i w_i$ 
adds the background image without ever scaling it down by coverage, which on a 
representative frame saturates $13.6\%$ of the board footprint to pure white
and renders the footprint $9\%$ larger than the true silhouette --- structure
a printed board cannot produce, and which an optimizer will nonetheless learn to use.
We process the artifact's rasterized primitives in chunks of $4{,}000$ to
keep peak GPU memory bounded during the EOT-augmented gradient pass.
The resulting attacked inputs $(I^{\mathcal{A}}, P^{\mathcal{A}})$ are fed 
into the calibration surrogate, and the gradient with respect to $C_{\mathcal{A}}$
flows back through both the calibration network and the rasterizer.

\vspace{-4mm}
\subsection{Training Recipe}
\label{sec:impl-training}

Universal-artifact training (Section~\ref{sec:attack-design}) is run on a 
16-frame training window of a single KITTI seq~13 segment, with a held-out
4-frame validation split.
At each optimization step we sample a mini-batch of $B = 4$ training frames; 
for each frame we sample $K_{\mathrm{eot}} = 8$ random pose perturbations 
$\delta T$ within $\pm$\,a few degrees rotation and $\pm 0.25$\,m translation 
around $T_{\mathcal{A}}$.
The corruption loss $\mathcal{L}_{\mathrm{corr}}$ (Eq.~\ref{eq:corrupt}) is 
computed on each EOT sample and averaged.
The trigger term $\mathcal{L}_{\mathrm{trig}}$ (Eq.~\ref{eq:trigger}) 
\emph{does} enter the texture gradient: it is backpropagated through the 
miscalibration detector during texture optimization rather than verified
post~hoc, because the artifact's fixed geometry alone does not fire the detector.
We hinge it at a margin of $10$ in the detector's logit units so that it 
stops consuming texture capacity once the detector already fires by that margin.
We optimize $C_{\mathcal{A}}$ with AdamW + OneCycleLR over $S = 2000$ steps, 
with the peak LR set to $3\times$ the base LR (typical base $5\times 10^{-2}$) 
and OneCycleLR's warmup fraction set to $0.1$.
The window width and the learning rate must be set together: a 16-frame run at
this learning rate overfits, with held-out validation damage falling monotonically
from $+53^\circ$ to $+30^\circ$ while the training objective keeps rising. 
A 48-frame window at a base LR of $2\times 10^{-3}$ removes the effect and is 
the recommended setting.
Validation is run every 50 steps with EOT disabled and the perturbation
$\delta T$ frozen from a cache to ensure deterministic comparison.
The best-on-validation artifact is saved at the end of training.
\vspace{-2mm}
\section{Evaluation}
\label{sec:evaluation}

We evaluate \name\ in three settings: on open autonomous-driving benchmarks, on a physical robot, and in a closed-loop driving simulator.
First, we evaluate the complete attack chain on public benchmarks.
The miscalibration detector triggers online calibration (Stage~1, Section~\ref{sec:eval-stage1}).
The artifact then drives the calibration toward an incorrect extrinsic (Stage~2, Section~\ref{sec:eval-stage2}).
This corrupted extrinsic degrades downstream perception tasks (Section~\ref{sec:eval-downstream}).
We establish the attack chain on KITTI and reproduce the same mechanism on nuScenes (Section~\ref{sec:eval-nuscenes}).
Next, we evaluate \name\ on a physical Husky robot (Section~\ref{sec:eval-husky}).
Finally, we demonstrate an end-to-end closed-loop attack in CARLA (Section~\ref{sec:eval-sim}).
These experiments validate the same corruption mechanism on real hardware and against a live driving stack.

\vspace{-2mm}
\subsection{Experimental Setup}
\label{sec:eval-setup}

\vspace{1mm}
\noindent\textbf{Datasets.}
Our primary benchmark is the KITTI odometry split with its synchronized RGB images, Velodyne HDL-64E point clouds, and factory-calibrated ground-truth extrinsics $T^\star_{\ell \rightarrow c}$.
Training of the universal artifact uses a small consecutive window of seq~13.
Deployment evaluation uses (a) all 3{,}281 frames of seq~13 in 13 disjoint 16-frame windows, and (b) four held-out test sequences (seq~14, 15, 20, 21).
We repeat the entire attack chain on the nuScenes benchmark.

\vspace{1mm}
\noindent\textbf{Platforms.}
Our physical-deployment target is a Clearpath Husky\,A300 mobile
robot~\cite{clearpath_huskya300} equipped with Luxonis OAK-D Pro camera~\cite{oakd_pro} and an Ouster OS1-128
LiDAR~\cite{ouster_os1}; all perception runs on the onboard NVIDIA
Jetson AGX Orin with the BEVFusion backbone compiled to
TensorRT\,int8.
The closed-loop evaluation runs in the Carla simulator.

\begin{figure}[t]
\centering
\includegraphics[width=.45\columnwidth]{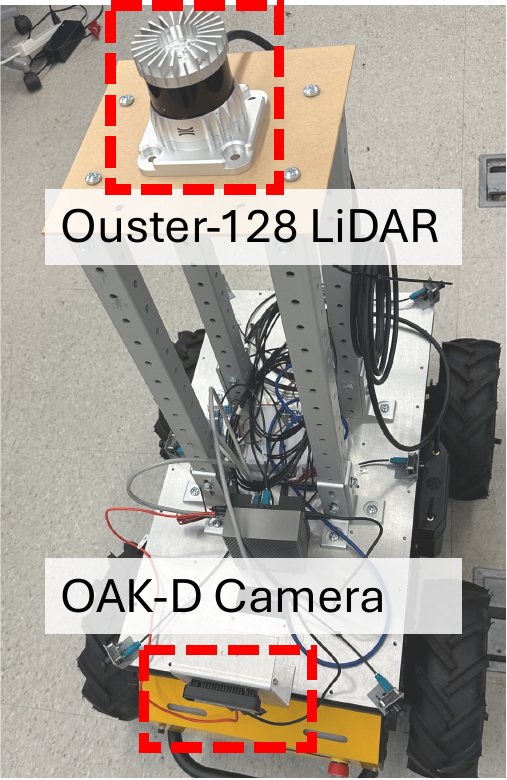}
\vspace{-2mm}
\caption{Deployment platform: Clearpath Husky\,A300 with Ouster
OS1-128 LiDAR and OAK-D depth camera, running \name{} on an NVIDIA Jetson AGX Orin.}
\label{fig:husky}
\vspace{-2mm}
\end{figure}

\vspace{1mm}
\noindent\textbf{Surrogate and target calibrators.}
The white-box matched surrogate is the surrogate-diffusion 
calibrator~\cite{jing2024surrogatediffusion} with a CalibNet 
backbone~\cite{iyer2018calibnet}; the diffusion wrapper and 
its backbone are one model, and we will refer to it as CalibNet.
Transferability is measured against LCCNet~\cite{lv2021lccnet}
and RGGNet~\cite{yuan2020rggnet}, each evaluated using its 
publicly available pretrained KITTI checkpoint.


\vspace{1mm}
\noindent\textbf{Baselines.}
We compare \name\ against two baselines: (i) the clean prediction 
without artifact; (ii) a \emph{gray board} of identical geometry, 
point count, and pose, carrying no adversarial content, which is 
the control that separates the attack from the board's mere 
physical presence. 

\vspace{1mm}
\noindent\textbf{Metrics.}
For each evaluation frame, we sample an initial miscalibration $T^{(0)}_{\ell \rightarrow c}$ at a fixed rotation/translation magnitude, run the calibrator's $K=10$ refinement iterations, and report:
\begin{itemize}[itemsep=0pt, topsep=2pt, parsep=0pt]
    \item \textbf{Rotation damage} $\Delta_\mathrm{rot} = \|\log(\Delta R)\|_2$ in degrees, where $\Delta R$ is the rotation component of $(T^\star_{\ell \rightarrow c})^{-1} \hat{T}_{\ell \rightarrow c}$. We report the \emph{marginal} damage --- the rotation error the artifact \emph{adds} relative to the clean (no-artifact) baseline on the same frame --- so as to isolate the attack's effect from the calibrator's residual error.
    \item \textbf{Translation damage} $\Delta_\mathrm{trans} = \|\Delta t\|_2$ in meters, defined analogously.
    \item \textbf{Attack success rate above $5^\circ$} (ASR$>5^\circ$), the fraction of evaluation frames on which $\Delta_\mathrm{rot} > 5^\circ$.
\end{itemize}
The $5^\circ$ threshold is chosen because it is the upper bound of the calibrator's nominal in-distribution residual error; rotation errors above $5^\circ$ are unambiguously attack-induced and produce visible misalignment in downstream camera--LiDAR projection.

\subsection{Stage 1: Triggering Online Calibration}
\label{sec:eval-stage1}

A complete attack begins with Stage~1: the miscalibration detector 
$\mathcal{D}$ must fire when the artifact is present, so that 
online calibration is invoked at all (Eq.~\ref{eq:goal-trigger}).
We validate this precondition before turning to the corruption 
it enables (Section~\ref{sec:eval-stage2}).



\vspace{1mm}
\noindent\textbf{Stage 1 on a global rigid-displacement detector.}
We evaluate the LiDAR--camera self-check detector of Wei {\em et al.}
\cite{wei2024online}, which scores the global rigid displacement between 
the two modalities. The same $3.6 \times 1.8$\,m board used throughout --- with 
\emph{no geometry change} --- triggers it as an attack should
(Table~\ref{tab:stage1-trigger}).

The texture is worth $+0.70$ of fire rate over its own gray control 
($0.28 \rightarrow 0.98$).
The effect is a conjunction rather than a sum: image-only is $0.00$ and returns-only 
is $0.00$, so the texture bites only when the board is also physically present 
in the depth channel.
Clean scenes never fire ($0.00$ across all held-out frames).
Importantly, an ordinary near-field object does not fire it either: a truck rear
face with \emph{more} returns and the \emph{same} occlusion gives $0.00$, 
so firing is not a size effect.
The trigger is also not a high-frequency artifact --- it withstands a $20$\,mm 
Gaussian ink smear on a $3.6$\,m board, collapsing only at $40$\,mm.

\begin{table}[t]
\centering
\small
\setlength{\tabcolsep}{5pt}
\begin{tabular}{lrr}
\toprule
condition ($96$ held-out frames) & fires & mean conf. \\
\midrule
\name, Stage-1-only texture   & $\mathbf{0.98}$ & $0.047$ \\
\name, \textbf{joint artifact} & $\mathbf{0.89}$ & $0.187$ \\
\bottomrule
\end{tabular}
\caption{Stage-1 firing against the global rigid-displacement detector of 
\cite{wei2024online}, at the same $3.6 \times 1.8$\,m board. \emph{fires} is 
the fraction of held-out frames flagged as miscalibrated. The gray-board ($0.28$)
and truck-rear ($0.00$) controls reported in the text that distinguishes an 
adversarial trigger from a near-field occlusion effect; both are passed.}
\label{tab:stage1-trigger}
\end{table}

Local RGB–depth consistency detectors do not defend against Stage 1 either, 
albeit for a different reason. Against the detector of Tahiraj 
{\em et al.}~\cite{tahiraj2025calornocal}, the firing is texture-blind:
the optimized poster and a gray board of identical geometry both fire it 
at $0.54$ on held-out frames, and the image channel alone contributes 
nothing ($0.00$). The detector reacts only to near-field occlusion, 
and to any of it --- an ordinary truck rear face at the same range fires 
it harder than the attack board ($0.95$), a bus harder still ($0.98$). 
Such a detector cannot shield the calibration module: it cannot separate 
the adversarial board from the board's mere presence, or from a passing-by 
vehicle, so it flags miscalibration and requests exactly the recalibration 
that Stage~2 then corrupts. 
The texture-driven trigger of the above 
paragraph is thus specific to global rigid-displacement self-checks; 
against a consistency detector the board triggers recalibration by 
presence rather than by optimization, and no benign near-field geometry
is exempted.

\vspace{1mm}
\noindent\textbf{Composing the two stages.}
Separately optimized textures do \emph{not} compose: a Stage-2 poster fires the 
detector at only $0.39$, and a Stage-1-targeted texture contributes $+0.05^\circ$ 
to the calibrator.
On this detector the trigger costs \emph{texture}, the same resource Stage~2 
consumes, so the two stages compete for one budget.
Optimizing a single texture against both objectives closes the gap, provided 
the Stage-1 term is \emph{hinged}: an unbounded margin keeps rewarding an ever
more negative detector score long after the detector already fires, decaying 
Stage-2 damage from $+43^\circ$ to $+8.6^\circ$ over $1000$ steps, whereas 
capping the term once firing is comfortable sends the remaining gradient to Stage~2.
The resulting joint artifact fires at $0.89$ and slightly \emph{exceeds} the 
Stage-2-only poster on every Stage-2 metric ($+14.79^\circ$ vs $+14.5^\circ$ at
$n=30$; $97\%$ vs $96\%$ ASR$>5^\circ$), so composition is effectively free.


\subsection{Stage 2: Corrupting the Extrinsic}
\label{sec:eval-stage2}


Once Stage~1 has triggered online calibration, the artifact's task is to drive
that calibration to a wrong extrinsic; this is our central result.
We report the headline physical-billboard attack across the calibrator's operating 
regime (Section~\ref{sec:eval-main}), isolate the cross-modal mechanism that 
produces it (Section~\ref{sec:eval-crossmodal}), and establish that one trained poster 
deploys across an entire driving sequence (Section~\ref{sec:eval-generalization}).

\vspace{-2mm}
\subsubsection{Physical Billboard Attack}
\label{sec:eval-main}

\vspace{1mm}
\noindent\textbf{Headline result.}
At $15^\circ$ initial miscalibration on KITTI seq~13, the deployed artifact adds 
$+18.9^\circ$ of marginal rotation damage at $96\%$ attack success rate above
$5^\circ$ ($n=200$ frames), against a $+3.5^\circ$ gray-board control of 
identical geometry.
Across the calibrator's entire trained operating regime the attack is essentially flat: 
$+19.4^\circ$, $+19.9^\circ$ and $+17.1^\circ$ at $5^\circ$, $10^\circ$ and $15^\circ$ 
initial miscalibration, at $96$--$98\%$ success (Table~\ref{tab:main-pert}) --- so 
it is not exploiting an out-of-distribution weakness of the calibrator.
Table~\ref{tab:main-pert} additionally reports the artifact under both print modes 
at $n=30$ to expose the print-mode spread; we report large-$n$ figures in text because 
a $30$-frame draw of this evaluation varies by up to $1.6\times$ in either direction.
Translation damage is small ($\leq 0.3$\,m, typically $\leq 0.1$\,m); a sweep over
the translation weight $\lambda_t$ in the corruption objective does not change 
this and we therefore treat \name\ as a rotation-only attack.

\begin{table}[t]
\centering
\small
\setlength{\tabcolsep}{3.5pt}
\begin{tabular}{lrrrr}
\toprule
init.\ miscal. & baseline & hard print & smooth print & gray ctrl \\
\midrule
$5^\circ$ (in-dist.)  & $2.2^\circ$ & $+10.8^\circ$ / $76\%$ & $+16.2^\circ$ / $96\%$ & --- \\
$10^\circ$ (in-dist.) & $3.0^\circ$ & $+10.3^\circ$ / $72\%$ & $+15.8^\circ$ / $92\%$ & --- \\
$15^\circ$ (in-dist.) & $2.3^\circ$ & $+15.6^\circ$ / $72\%$ & $+27.1^\circ$ / $100\%$ & $+3.5^\circ$ \\
$25^\circ$ (OOD)      & $6.6^\circ$ & {$+17.9^\circ$ / $64\%$} & $+29.0^\circ$ / $92\%$ & --- \\
\bottomrule
\end{tabular}
\caption{\name\ vs.\ initial miscalibration magnitude on KITTI. Cells are marginal 
rotation damage / ASR$>\!5^\circ$. The trained calibrator's in-distribution range 
is $5$--$15^\circ$.}
\label{tab:main-pert}
\end{table}

\vspace{1mm}
\noindent\textbf{Cross-sequence universality.}
A poster trained on \emph{a single small window of seq~13} generalizes without 
retraining to four held-out test sequences (Table~\ref{tab:cross-seq}).
Aggregated over $150$ frames spanning all held-out sequences, the attack induces 
$+16.2^\circ$ ($76\%$ ASR$>5^\circ$) under a hard-patch print and $+33.9^\circ$
($92\%$) under a smooth print, against a gray-board control of $+3.3^\circ$ ($18\%$).
The attack is therefore $5$--$10\times$ its own occlusion floor on sequences
the poster has never seen.
This is the strongest evidence that a single adversarial billboard, once optimized, 
can be deployed broadly across an AV's operating distribution.

\begin{table}[t]
\centering
\small
\begin{tabular}{lrr}
\toprule
condition (held-out seqs, $n=150$) & $\Delta_\mathrm{rot}$ & ASR$>5^\circ$ \\
\midrule
gray board (occlusion control) & $+3.3^\circ$ & $18\%$ \\
\name, hard-patch print       &$+16.2^\circ$ & $76\%$ \\
\name, smooth print           & $+33.9^\circ$ & $92\%$ \\
\bottomrule
\end{tabular}
\caption{Cross-sequence universality, aggregated over held-out seq~14,
15, 20 and 21. 
The poster is trained on one window of seq~13 and evaluated unchanged.}
\label{tab:cross-seq}
\end{table}

\vspace{-2mm}
\subsubsection{Cross-Modal Mechanism}
\label{sec:eval-crossmodal}
\label{sec:eval-physical}

To verify that \name\ is genuinely a cross-modal attack rather than a 
single-modality artifact, we isolate the two channels the billboard 
injects into.
The decisive comparison is the one already present in every table above: 
a gray board of identical geometry, point count, and pose --- the full 
LiDAR and image footprint with the adversarial content removed --- produces 
$+3.3$ to $+3.8^\circ$, while the same board carrying the optimized texture 
produces $+14.5$ to $+18.9^\circ$. The texture is therefore worth $+8.6$ to 
$+8.9^\circ$ over the board's mere physical presence. Neither channel 
accounts for the attack on its own: the texture is inert without the 
board's LiDAR returns to anchor it, and the returns alone sit at the 
occlusion floor. The two together engineer a false correspondence between
the image pattern and the LiDAR returns, which the calibrator reads as 
evidence of a specific, incorrect extrinsic. As shown in 
Section~\ref{sec:eval-stage1}, the same conjunction holds for the 
Stage-1 detector, where the image and depth channels in isolation 
both score exactly the clean-scene rate.

\subsubsection{Training Size and Deployment Universality}
\label{sec:eval-generalization}

\vspace{1mm}
\noindent\textbf{Universal training on a small frame set.}
\label{sec:universal-training-size}
The universal artifact is trained on a \emph{focused} window --- a short 
run of consecutive frames from a single sequence --- rather than on 
a wide, diverse sample of the dataset: a diverse training set averages 
the per-frame gradients into a mild poster, while a scene-consistent 
window lets the optimizer commit to an extreme texture that still 
generalizes to unseen sequences (Table~\ref{tab:cross-seq}).
How narrow that window can be is set by overfitting. At $16$ frames and
a learning rate of $8\times10^{-3}$, held-out damage falls monotonically 
from $+53^\circ$ to $+30^\circ$ while the training objective rises; 
widening to $48$ frames at $2\times10^{-3}$ removes the divergence and
yields our strongest artifact.
We select checkpoints on held-out validation rather than on 
the training objective throughout.

\label{sec:deployment-universality}
\vspace{1mm}
\noindent\textbf{Deployment across all of seq~13.}
We evaluate the trained artifact, with no further training, on 13 
disjoint 16-frame windows spanning all 3{,}281 frames of seq~13,
pairing each window with its own gray-board control so that a 
window's damage is separated from the board's mere occlusion 
(Table~\ref{tab:windows}).
Twelve of the 13 windows ignite --- mean marginal damage at least 
$5^\circ$ above that window's own control --- for an aggregate 
$+22.1^\circ$ at $94\%$ ASR$>5^\circ$ under a smooth print, 
against a $+3.2^\circ$ aggregate control.
The one exception (window 2000--2015, $+4.7^\circ$ against 
its own $+4.9^\circ$ control) and the $13.5\times$ spread of
per-window strength track the scene rather than the poster, so a 
trained artifact deploys broadly but an attacker cannot assume 
that every roadside location is equally exploitable.

\begin{table}[t]
\centering
\small
\setlength{\tabcolsep}{4pt}
\begin{tabular}{lrrrrr}
\toprule
& {gray} & \multicolumn{2}{c}{{hard print}} & \multicolumn{2}{c}{{smooth print}} \\
\cmidrule(lr){3-4}\cmidrule(lr){5-6}
{window (seq 13)} & {$\Delta_\mathrm{rot}$} & {$\Delta_\mathrm{rot}$} & {ASR} & {$\Delta_\mathrm{rot}$} & {ASR} \\
\midrule
{0--15}       & {$+1.3^\circ$}  & {$+8.6^\circ$}  & {$100\%$} & {$+11.3^\circ$} & {$100\%$} \\
{50--65}      & {$+3.0^\circ$}  & {$+5.6^\circ$}  & {$62\%$}  & {$+10.4^\circ$} & {$100\%$} \\
{100--115}    & {$+1.5^\circ$}  & {$+6.7^\circ$}  & {$100\%$} & {$+9.3^\circ$}  & {$100\%$} \\
{200--215}    & {$+3.1^\circ$}  & {$+11.9^\circ$} & {$100\%$} & {$+15.7^\circ$} & {$100\%$} \\
{500--515}    & {$+3.7^\circ$}  & {$+26.3^\circ$} & {$94\%$}  & {$+57.9^\circ$} & {$100\%$} \\
{800--815}    & {$+1.5^\circ$}  & {$+7.8^\circ$}  & {$100\%$} & {$+13.4^\circ$} & {$100\%$} \\
{1100--1115}  & {$+1.4^\circ$}  & {$+5.7^\circ$}  & {$94\%$}  & {$+8.7^\circ$}  & {$100\%$} \\
{1400--1415}  & {$+10.9^\circ$} & {$+22.1^\circ$} & {$50\%$}  & {$+60.8^\circ$} & {$88\%$} \\
{1700--1715}  & {$+2.5^\circ$}  & {$+4.9^\circ$}  & {$44\%$}  & {$+8.3^\circ$}  & {$100\%$} \\
{2000--2015}  & {$+4.9^\circ$}  & {$+2.6^\circ$}  & {$0\%$}   & {$+4.7^\circ$}  & {$31\%$} \\
{2300--2315}  & {$+3.3^\circ$}  & {$+19.2^\circ$} & {$81\%$}  & {$+63.5^\circ$} & {$100\%$} \\
{2600--2615}  & {$+1.0^\circ$}  & {$+7.7^\circ$}  & {$100\%$} & {$+11.7^\circ$} & {$100\%$} \\
{2900--2915}  & {$+3.5^\circ$}  & {$+10.1^\circ$} & {$100\%$} & {$+12.1^\circ$} & {$100\%$} \\
\midrule
{all ($n=208$)} & {$+3.2^\circ$} & {$+10.7^\circ$} & {$79\%$} & {$\mathbf{+22.1^\circ}$} & {$\mathbf{94\%}$} \\
\bottomrule
\end{tabular}
\caption{{Cross-window deployment of one trained artifact across 
all of KITTI seq~13, at $15^\circ$ initial miscalibration. Cells are 
marginal rotation damage and ASR$>\!5^\circ$; \emph{gray} is the same 
board with the adversarial content removed, measured per window.
Twelve of 13 windows exceed their own control by more than $5^\circ$;
window 2000--2015 does not.}}
\label{tab:windows}
\end{table}




\vspace{1mm}
\noindent\textbf{Surrogate scope and transferability}.
\label{sec:transferability}
We optimize \name against CalibNet and evaluate the same artifact on two held-out calibration architectures without re-optimization. As shown in Table~\ref{tab:transfer}, \name induces a $+18.87^\circ$ error on its white-box surrogate, substantially exceeding the gray and scrambled controls. On the black-box LCCNet model, \name still produces a $+6.55^\circ$ error, compared with $+0.98^\circ$ for the gray board, demonstrating partial cross-architecture transfer. However, the scrambled control causes an even larger $+8.87^\circ$ error, suggesting that LCCNet is also broadly sensitive to patterned visual occlusion and making the transfer benefit less specific to the optimized texture. On RGGNet, \name produces only a $+1.16^\circ$ error, comparable to both controls. Overall, the attack exhibits meaningful black-box transfer to LCCNet, but its transferability is architecture-dependent and does not extend to RGGNet.

\begin{table}[t]
\centering
\small
\setlength{\tabcolsep}{4pt}
\begin{tabular}{llrrr}
\toprule
target & role & {gray} & {scram.} & {\name} \\
\midrule
CalibNet~\cite{iyer2018calibnet} & {white-box} & {$+3.43^\circ$} & {$+3.38^\circ$} & {$+18.87^\circ$} \\
LCCNet~\cite{lv2021lccnet}   & transfer & {$+0.98^\circ$} & {$+8.87^\circ$} & {$+6.55^\circ$} \\
RGGNet~\cite{yuan2020rggnet} & transfer & {$+1.44^\circ$} & {$+1.21^\circ$} & {$+1.16^\circ$} \\
\bottomrule
\end{tabular}
\caption{{Cross-architecture evaluation of an artifact optimized against CalibNet. LCCNet and RGGNet are evaluated without re-optimization.
}}
\label{tab:transfer}
\end{table}

\vspace{-2mm}
\subsection{Downstream Consequences}
\label{sec:eval-downstream}


A corrupted $T_{\ell \rightarrow c}$ does not merely inflate an intrinsic 
calibration metric; it deterministically misaligns every downstream 
camera--LiDAR fusion operation, because each is parameterized by the
same extrinsic.
We measure this end-to-end effect on 3D object detection, the canonical
fusion task, using two detectors on the KITTI object validation split
(3{,}769 frames): PointPainting~\cite{vora2020pointpainting}, 
a camera--LiDAR fusion detector that paints each LiDAR point with 
the image semantics at its projected pixel, and 
PointPillars~\cite{lang2019pointpillars}, a LiDAR-only detector 
that never reads the extrinsic.
Each is run on the clean scene and on the attacked scene, 
where the attacked scene adds the billboard's {realizable LiDAR return} 
and supplies the corrupted extrinsic the calibrator converged to under 
the artifact {(mean marginal rotation damage $+58.1^\circ$, 
median $+60.2^\circ$, with $98\%$ of frames above $10^\circ$)}.
PointPillars is a control: it receives the same artifact points but,
being extrinsic-free, isolates whether any detection damage comes 
from the physical points or from the calibration corruption.

Table~\ref{tab:downstream} reports Car 3D AP@R40 (Moderate, IoU $0.7$).
{Under \name, PointPainting collapses from $51.72$ to $0.36$ AP --- a
$99.3\%$ loss --- while PointPillars is statistically unchanged
($78.41 \rightarrow 78.45$, $\Delta\!=\!+0.04$).}
The control is decisive: the {artifact's} points alone do nothing
to a detector that does not consume the extrinsic, so the fusion 
collapse cannot be attributed to the physical artifact and must be 
the calibration corruption.
{The $+58^\circ$ realized on this split exceeds the $+18.9^\circ$
headline because the object split is a different scene distribution 
than the odometry windows, and per-scene strength varies more than tenfold 
(Section~\ref{sec:deployment-universality}); we report the odometry figure 
as the headline because it is the more conservative of the two, and
the downstream conclusion follows from either --- fusion has already 
broken by $\sim\!10^\circ$, and $98\%$ of these $3{,}769$ frames 
carry more than that.}

\begin{table}[t]
\centering
\small
\begin{tabular}{llrr}
\toprule
detector & input & clean & attacked \\
\midrule
PointPainting (fusion)  & cam+LiDAR  & $51.72$ & {$\mathbf{0.36}$} \\
PointPillars (control)  & LiDAR-only & $78.45$ & {$78.45$} \\
\bottomrule
\end{tabular}
\caption{End-to-end downstream impact: Car 3D AP@R40 (Moderate, IoU $0.7$) on KITTI val (3{,}769 frames). The fusion detector collapses under \name; the extrinsic-free LiDAR-only control is unchanged ({$\Delta\!=\!+0.04$}), isolating the calibration corruption as the cause.}
\label{tab:downstream}
\end{table}

\vspace{1mm}
\noindent\textbf{Wrong cross-modal association alone is near-fatal.}
A deployed fusion detector uses the single corrupted extrinsic for two 
operations: the field-of-view (FoV) crop that selects which LiDAR points 
enter the detector, and the projection lookup that paints each point with
image semantics.
We separate them (Table~\ref{tab:downstream-mech}).
Design~B (the deployment setting) corrupts both: under the rotated 
extrinsic the real cars project off the image and are dropped from the 
painted cloud, and the surviving points carry wrong semantics; 
{AP is $0.36$}.
Design~A instead uses the ground-truth extrinsic for the FoV 
crop --- so the real cars remain in the cloud with correct geometry --- and 
corrupts only the paint lookup; {AP is still $1.51$}, almost as severe.
Corrupting the cross-modal \emph{association} alone is therefore sufficient 
to break fusion even with perfect geometry; \name\ hits fusion through both 
pathways, and either alone is near-fatal.
{(Design~A additionally feeds PointPainting in-range background-painted 
points that are partly out of its training distribution --- itself 
a genuine consequence of the attack, since a detector handed a badly 
rotated extrinsic does paint front cars as background.)}

\begin{table}[t]
\centering
\small
\begin{tabular}{lrl}
\toprule
PointPainting condition & AP & corrupted \\
\midrule
clean                          & $51.72$ & nothing \\
Design~A (GT FoV, bad paint)   & {$1.51$}  & association only \\
\textbf{Design~B} (bad FoV+paint) & {$\mathbf{0.36}$} & association + FoV \\
\bottomrule
\end{tabular}
\caption{Mechanism ablation. Corrupting only the cross-modal association 
(Design~A, geometry correct) is near-fatal on its own; the full
deployment (Design~B) additionally drops the real cars from the FoV crop.}
\label{tab:downstream-mech}
\end{table}

{These results are not a single-model artifact: because every fusion 
operation is parameterized by $T_{\ell \rightarrow c}$ deterministically, 
the same corruption propagates verbatim into BEV occupancy, depth-aware 
segmentation, and sensor-fused tracking --- which Section~\ref{sec:eval-sim} 
confirms end-to-end against a live driving stack.}

\vspace{-2mm}
\subsection{{Generalization to nuScenes}}
\label{sec:eval-nuscenes}

{To test that the attack chain is a property of the calibration mechanism 
rather than of one dataset, we repeat it on nuScenes, whose sensor suite
and LiDAR frame convention both differ from KITTI.
We fine-tune the CalibNet surrogate on nuScenes v1.0-trainval 
from the KITTI checkpoint and run the same universal poster attack.}

\vspace{1mm}
\noindent\textbf{{Attack.}}
{At $15^\circ$ initial miscalibration, the deployed poster adds
$+39$--$47^\circ$ of raw rotation damage at $89$--$98\%$ success above 
$5^\circ$, against a gray-board control of $+15$--$22^\circ$; the net 
texture effect (poster minus gray) is $+29^\circ$ near the training window 
and $+18^\circ$ on held-out cross-scene frames (Table~\ref{tab:nusc-attack}), 
inside the $+16$--$34^\circ$ band KITTI reaches under a comparable renderer.
Unlike on KITTI, the attack additionally moves translation by up to $+0.5$\,m.
The higher gray-board baseline reflects a weaker calibrator: the nuScenes 
surrogate corrects only about half of a $15^\circ$ miscalibration against 
KITTI's $\sim\!85\%$, so a near-field board occludes more of the 
recoverable signal.
One dataset-specific detail is worth recording: the billboard's default 
pose assumes KITTI's $x$-forward LiDAR, and nuScenes LIDAR\_TOP is 
$y$-forward, so the artifact pose must be derived from the extrinsic --- placed 
a fixed distance ahead of the camera and facing it --- which is correct for 
either convention and leaves the KITTI results unchanged.}

\vspace{1mm}
\noindent\textbf{Downstream.}
The corrupted extrinsic breaks camera--LiDAR fusion here as it does on KITTI, measured directly against the dataset's 3D boxes.
Projecting each nuScenes ground-truth box to the front camera through the deployed extrinsic ($902$ object views over $50$ validation frames), the poster drives the mean box IoU from $0.38$ under clean calibration to near zero: of the objects the clean calibration localizes correctly (IoU $>0.5$), the attack destroys $96\%$ (IoU $<0.1$), against $86\%$ for the gray-board control.
Equivalently, each LiDAR point's camera-feature lookup is displaced a median $437$\,px in a $512$\,px-wide image, so essentially every object's cross-modal features are misassociated --- the same failure that collapses KITTI fusion AP.
This evaluation uses the same splat renderer and full point cloud as the KITTI $+30$--$70^\circ$ figures rather than the faithful renderer of the $+16$--$34^\circ$ band, so both the poster and its control are inflated by dense near-field returns.

\begin{table}[t]
\centering
\small
\resizebox{\columnwidth}{!}{%
\begin{tabular}{lrrrr}
\toprule
{region ($15^\circ$ init)} & {poster} & {ASR$>5^\circ$} & {gray} & {net (mean/med)} \\
\midrule
{trained window}             & {$+44.1^\circ$} & {$89\%$} & {$+15.1^\circ$} & {$+29.0$/$+28.9^\circ$} \\
{near held-out (same scene)} & {$+47.2^\circ$} & {$98\%$} & {$+17.9^\circ$} & {$+29.3$/$+28.5^\circ$} \\
{far held-out (cross-scene)} & {$+39.4^\circ$} & {$94\%$} & {$+21.6^\circ$} & {$+17.8$/$+15.6^\circ$} \\
\bottomrule
\end{tabular}
}
\caption{{\name\ on nuScenes at $15^\circ$ initial miscalibration, with a CalibNet surrogate fine-tuned on v1.0-trainval. Cells are marginal rotation damage (poster) and ASR$>\!5^\circ$; \emph{gray} is the same board with no adversarial content; \emph{net} isolates the learned texture. The net effect matches KITTI's $+16$--$34^\circ$ band.}}
\label{tab:nusc-attack}
\end{table}

\subsection{Closed-loop Attack Simulation}
\label{sec:eval-sim}

\begin{figure}[t]
  \centering
  \begin{subfigure}[t]{0.23\textwidth}
    \includegraphics[width=\textwidth]{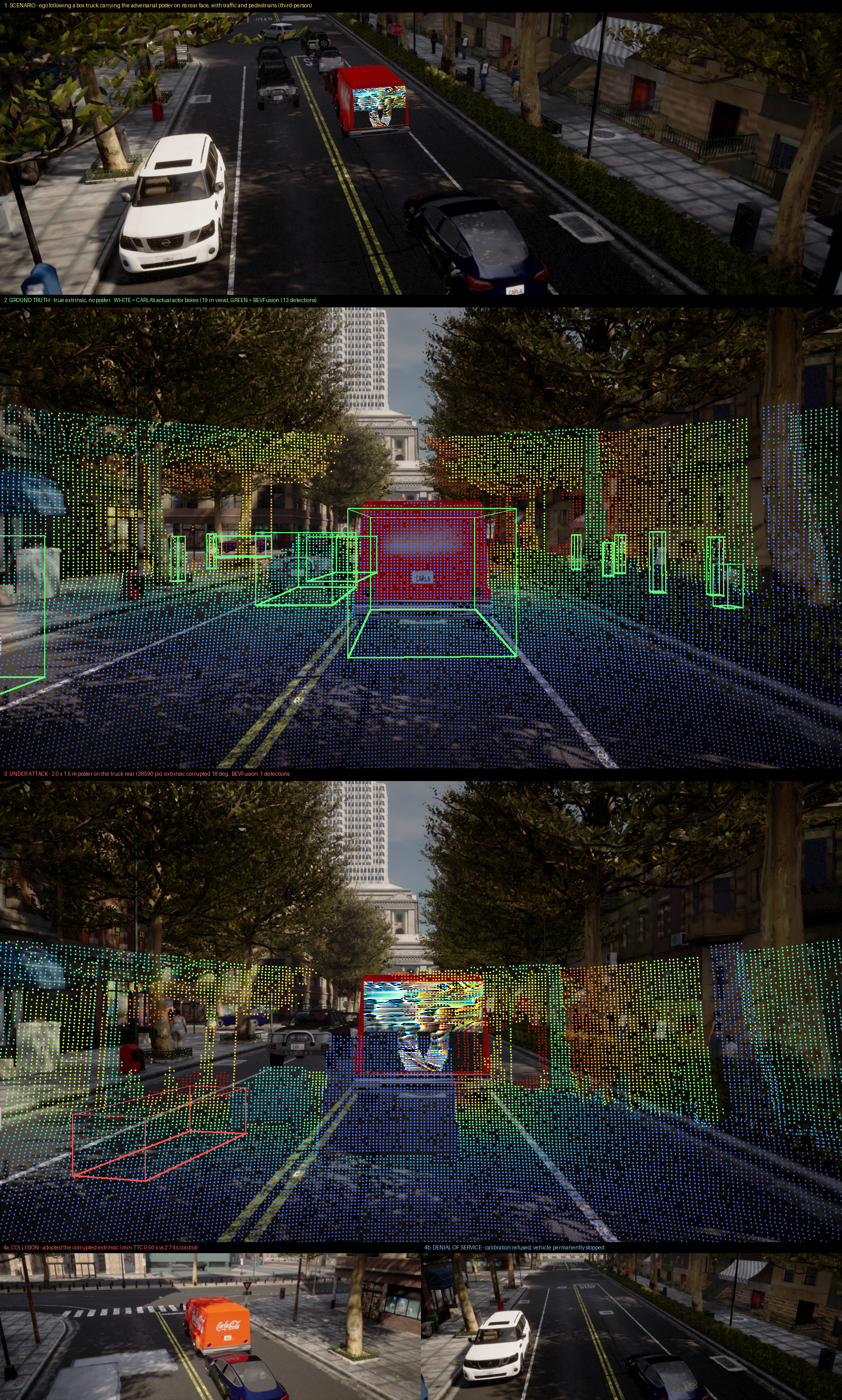}
    \caption{Poster on a lead truck's rear face.}
    \label{fig:truck}
  \end{subfigure}
  \hfill
  \begin{subfigure}[t]{0.23\textwidth}
    \includegraphics[width=\textwidth]{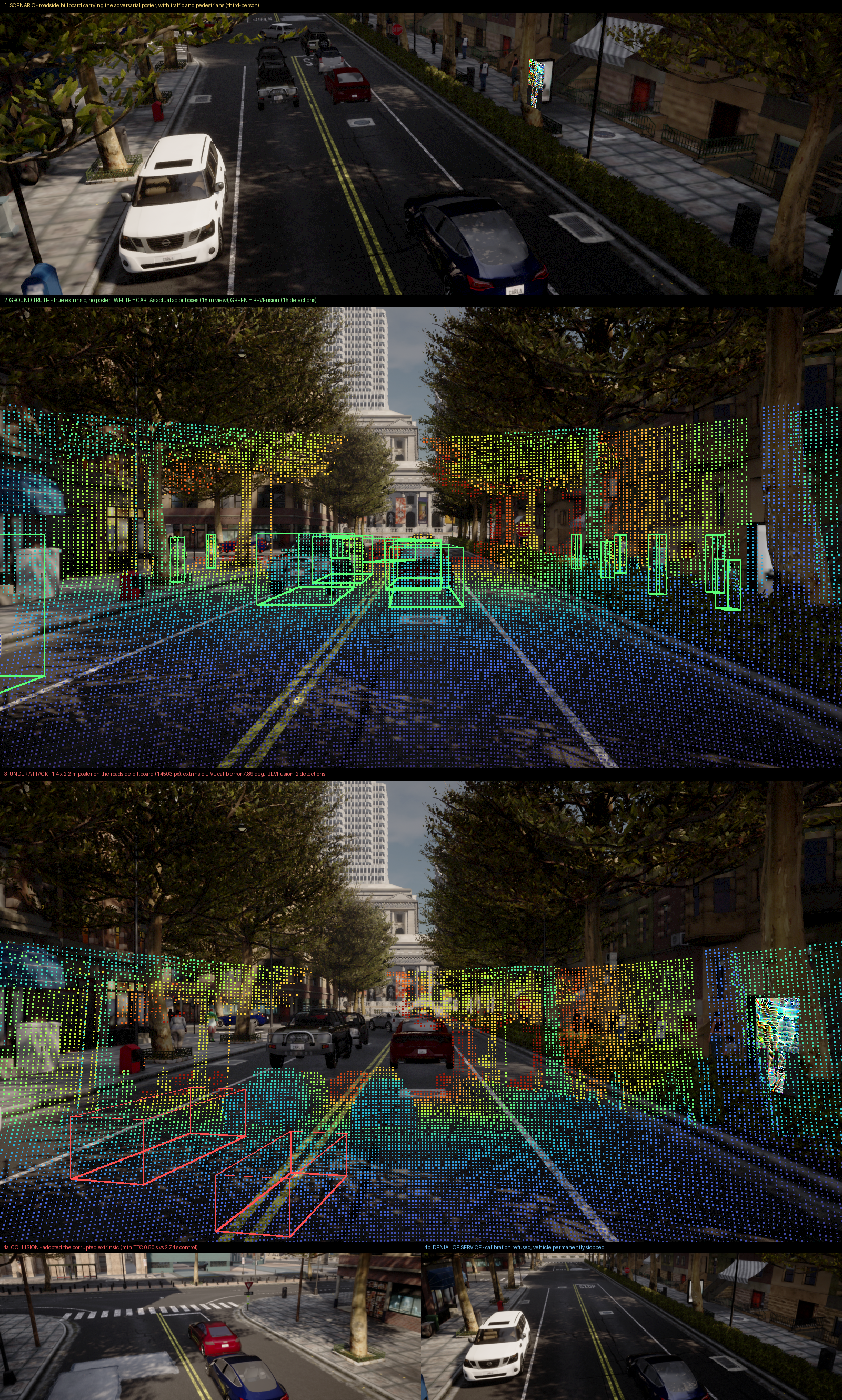}
    \caption{Poster on a roadside billboard.}
    \label{fig:billboard}
  \end{subfigure}
  \caption{End-to-end attack in closed-loop Carla simulation. Each column, top to bottom: (1) the scenario
  in third person; (2) clean scenario without attack (green
  BEVFusion detection boxes); (3) the attacked frame, where the online calibrator is triggered and poisoned, leading to miscalibration errors of $7.72^\circ$ and $7.89^\circ$, respectively, and wrong BEVFusion detection (red); and (4) the consequence---collision if adopting the miscalibration,
  permanent stop if triggering a failsafe policy.}
  \label{fig:endtoend}
\end{figure}

We evaluate the attack in Carla~0.9.16 (synchronous, 20\,Hz) with a six-camera ring
($1600\times900$, $80^\circ$ FoV) and a 128-beam LiDAR. The victim stack
comprises CalibNet for online camera--LiDAR calibration, BEVFusion for 3D detection, and a
pure-pursuit lateral controller with constant-time-gap adaptive cruise control, all finetuned on Carla data.

The artifact is placed two ways: a $2.0\times1.5$\,m poster on the rear face of a
box truck followed at 9\,m, and a $1.4\times2.2$\,m roadside billboard on the
sidewalk 13\,m ahead. Both are real objects in the simulator: their LiDAR
returns and occlusion come from Carla's ray cast, and only the printed texture is
optimized. The ego begins correctly calibrated; on approach, the online
calibrator is triggered and converges to an extrinsic $7.72^\circ$ (truck) and $7.89^\circ$
(billboard) from truth.

The downstream impact is what matters operationally. Under the corrupted
extrinsic, camera--LiDAR fusion is fed inconsistent geometry and BEVFusion
detections on the same frame collapse from 13 to 1 and from 15 to 2.
Objects that survive are displaced: a vehicle
reported 1.4\,m off is attributed to the wrong lane. The consequence then depends
on how the stack treats an unverifiable calibration result, and neither option is
safe. A trusting policy adopts the estimate, drives on misaligned fusion, and
\emph{collides}: with the artifact-carrying truck in the first scenario, and at
4.50\,m closest approach with a stopped vehicle in the second, with
time-to-collision falling from 2.74\,s to 0.50\,s. A conservative policy rejects
the implausible estimate, never reconverges while the artifact is in view, and
brings the vehicle to a \emph{permanent stop}---a denial of service on a live
carriageway. One roadside object therefore produces either a crash or an immobile
vehicle, and the attacker chooses neither outcome.

\vspace{-2mm}
\subsection{Real Deployment on the Husky Robot}
\label{sec:eval-husky}

To evaluate \name\ under real printing and sensing effects, we place
a printed adversarial board in the joint FoV of the OAK-D camera 
and Ouster OS1-128 LiDAR mounted on the Husky A300 described in 
Section~\ref{sec:eval-setup}. Thus, the camera observes the physically 
printed texture, while the LiDAR measures the board's actual geometry
and occlusion. We collect $3{,}391$ synchronized camera--LiDAR frames 
during a $22.6$-minute drive through outdoor plazas, parking lots, 
and indoor corridors.

The $3.6\times1.8$ m billboard used in the KITTI experiments is too 
large for this deployment. We therefore fabricate a fixed 
$1.14\times0.57$ m board that fits within the sensors' joint FoV 
at a $1.9$ m standoff. The board is printed at its designed dimensions 
without rescaling. We optimize its texture end-to-end using the 
rig's measured print-and-capture response and an 
expectation-over-transformation (EOT) loop that samples up to 
$\pm1.5^\circ$ of placement error. We then evaluate the printed 
board under perfect placement, $\pm1.5^\circ$ error, and a larger
unseen error of $\pm3^\circ$. A gray board with identical 
dimensions controls for geometry and occlusion effects.

\begin{figure}[h]
    \centering
    \includegraphics[width=\columnwidth]{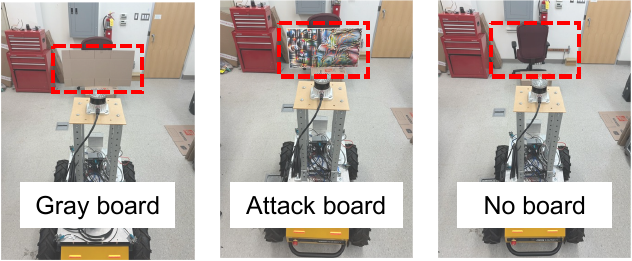}
    \vspace{-2mm}
    \caption{Three cases: gray board, attack board, and no board.}
    \vspace{-2mm}
    \label{fig:husky-attack}
\end{figure}

\begin{table}[t]
\centering
\small
\setlength{\tabcolsep}{3.5pt}
\begin{tabular}{@{}lrrr@{}}
\toprule
Placement error & Gray control & \name{} (mean/median)
& ASR$_{>5^\circ}$ \\
\midrule
$0^\circ$       & $-0.5^\circ$ & $+6.1^\circ / +10.0^\circ$ & $62\%$ \\
$\pm1.5^\circ$  & $-0.3^\circ$ & $+6.4^\circ / +9.8^\circ$  & $62\%$ \\
$\pm3^\circ$    & $-0.4^\circ$ & $+4.5^\circ / +7.1^\circ$  & $59\%$ \\
\bottomrule
\end{tabular}
\caption{Physical Husky evaluation. \name{} values show mean/median
marginal rotation damage over the gray control; ASR is the fraction
exceeding $5^\circ$. Sample counts are $n=120$, $840$, and $840$ by row.}
\label{tab:realrig}
\end{table}

As shown in Table~\ref{tab:realrig}, \name\ remains effective under 
physical placement errors. Within the EOT range of $\pm1.5^\circ$, 
it causes $+6.4^\circ$ mean and $+9.8^\circ$ median rotation damage, 
with a $62\%$ ASR$>5^\circ$, which is nearly identical to perfect placement.
Even at the unseen $\pm3^\circ$ error, it retains $+4.5^\circ$ mean
damage and a $59\%$ ASR. In contrast, the gray control remains 
within $0.5^\circ$ and has at most $1\%$ ASR, showing that the attack
is driven by the optimized texture rather than the board's geometry
or occlusion.

\vspace{-2mm}
\section{Discussion}
\label{sec:discussion}


\noindent\textbf{Potential defenses.}
Existing AV defenses can raise the difficulty of ACA, but do not directly
eliminate the calibration-layer vulnerability.
Cross-sensor consistency defenses~\cite{jia2020optical,yu2024physense,
xu2024physcout,ccs2024visionguard,man2023spatiotemporal} can detect
disagreement between modalities, but ACA uses a physical object that produces
consistent camera and LiDAR observations; moreover, our results show that
miscalibration detectors based on similar consistency signals can themselves
be triggered by adversarial or benign near-field objects~\cite{tahiraj2025calornocal,
wei2024online}.
Physical-invariant defenses~\cite{quinonez2020savior,nassi2020phantom,
sun2020lidar} and temporal validation~\cite{xiao2023wraith,kim2023adopt,
anon2021tempconsistency,anon2021shadowcatcher,hau2022shadows,
acsac2024intensity,sensys2024lidardefense,ccs2025realtimelidar,
xu2025hyper3def} could further reject implausible or unstable updates.
However, calibration drift is itself legitimate, making it difficult to set
tight rejection thresholds without preventing necessary recalibration.
Adversarial training~\cite{cheng2023depthhardening} may also improve robustness,
but would need to jointly protect both the miscalibration detector and
calibrator and may not generalize to unseen physical artifacts.
A more complete defense should therefore independently validate the proposed
extrinsic before committing it, using trusted historical bounds, independent
geometric evidence, or diverse calibration estimators, rather than relying
solely on the same camera--LiDAR observations that produced the update.

\vspace{1mm}
\noindent\textbf{Limitations and threats to validity.}
Our results establish the vulnerability of representative online-calibration
designs, but should not be interpreted as showing that all calibration systems
are equally vulnerable. Attack transfer is architecture-dependent, indicating
that model design can substantially affect susceptibility. More importantly,
production AV calibration pipelines and their update safeguards are largely
closed source, preventing direct evaluation against industry-level
implementations. Thus, the prevalence and severity of this vulnerability in
commercial AVs remain unknown. The attack also retains practical constraints,
including limited architecture-agnostic black-box transfer and sensitivity to
deployment geometry. These limitations bound the generality of our quantitative
results, while leaving the central finding unchanged: online calibration creates
a mutable, safety-critical state that requires explicit security protection.

\vspace{-3mm}
\section{Conclusion}
\label{sec:conclusion}

We presented \name, the first physical-world attack against 
camera--LiDAR online calibration, establishing calibration as a 
novel attack plane in the AV perception stack. Using a single 
adversarial poster, \name\ compromises both stages of the
calibration pipeline: it first spoofs the miscalibration detector
to trigger an unnecessary update and then steers the estimator toward 
an incorrect extrinsic transformation. Because this transformation is 
reused by subsequent fusion operations, one corrupted update creates 
persistent, system-wide misalignment. Across KITTI and nuScenes, 
\name\ substantially corrupts calibration and reduces PointPainting 
Car 3D AP@R40 from $51.72$ to $0.36$, while an extrinsic-free LiDAR-only 
detector remains unaffected. CARLA experiments further expose a 
safety--availability dilemma: accepting the corrupted estimate causes 
a collision, whereas rejecting it can leave the vehicle permanently 
stopped. Experiments with a printed board on a real Husky robot with 
camera--LiDAR further demonstrate the attack's physical-world 
feasibility.





\cleardoublepage
\appendix
\section*{Open Science}

\paragraph{Artifact availability.}
An anonymized artifact is available at \url{https://anonymous.4open.science/r/Calib-Attack-Artifact-1251}. The repository contains the attack implementation, configurations, evaluation scripts, environment specifications, and instructions needed to reproduce the paper's digital experiments. It also includes scripts for processing the KITTI and nuScenes datasets, the CARLA evaluation configuration, physical-experiment logs and videos, metric-computation code, and the data used to generate the paper's tables and figures. Random seeds, software versions, model configurations, and representative commands are documented in the accompanying README.

\paragraph{Third-party components.}
We do not redistribute datasets, pretrained models, simulator assets, or other components whose licenses prohibit redistribution. The artifact identifies the exact versions used and provides official download links and preprocessing instructions. Where permitted, we provide checksums or configuration information to help reviewers verify that the correct components are installed.

\paragraph{Safety-related limitation.}
The review artifact includes the high-resolution adversarial texture and the configuration required to evaluate the physical attack. Because public release of a print-ready texture and a fully automated physical-deployment pipeline could substantially reduce the effort required for misuse, we request that these components remain confidential to the review committee. Unless the disclosure process indicates that public release is safe, the post-acceptance public artifact will replace the texture with a downsampled, non-printable visualization and will omit the automated physical-deployment pipeline. All evaluation code, digital attack components, experimental results, and defense-related materials that do not create this additional physical risk will remain publicly available.

The review repository contains no author identities, institutional information, identifying commit history, or tracking mechanisms. If the paper is accepted, we will publish the releasable artifacts at a stable, non-anonymous URL and update this appendix accordingly.

\section*{Ethical Considerations}

\paragraph{Stakeholders and potential impacts.}
We considered the stakeholders who may be affected by this research, including vehicle occupants, pedestrians and other road users, autonomous-system developers, calibration-component maintainers, researchers, and society at large. Our attack exposes a safety-critical weakness in online camera--LiDAR calibration. Publication can help developers recognize calibration as an upstream attack surface and design appropriate safeguards. However, the technique is dual-use. A malicious actor could adapt it to disrupt the perception of a deployed autonomous system, potentially causing property damage, financial loss, or physical harm.

\paragraph{Research safeguards.}
We conducted digital experiments using simulation and established public autonomous-driving datasets. Physical experiments used equipment owned or controlled by the research team and were performed in a controlled environment without public traffic or uninvolved vehicles. The robotic platform remained under researcher supervision, with an immediate means to stop the experiment. We did not deploy the attack on public roads, access third-party systems, or interfere with operational vehicles or infrastructure. The study involved no recruited human participants and collected no private personal information. Public datasets were used under their applicable licenses, and we did not attempt to identify individuals appearing in them.

\paragraph{Disclosure and publication safeguards.}
Before public release, we will notify the maintainers of the evaluated calibration components and share the vulnerability mechanism and potential mitigations. To avoid unnecessarily lowering the barrier to physical misuse, the public artifact will not initially include the print-ready, high-resolution adversarial texture or a fully automated physical-deployment pipeline. These materials will be available to assigned reviewers for confidential evaluation. The public artifact will instead provide a non-printable visualization, evaluation code, experimental logs, and sufficient digital components to validate the paper's principal findings.

\paragraph{Decision and residual risk.}
Our analysis follows the principles of beneficence, respect for persons, justice, and respect for law and public interest. A residual risk remains that an adversary could reimplement the attack from the technical description or transfer the idea to other sensor combinations. Nevertheless, the attack requires physical placement within the sensors' shared field of view and knowledge of the targeted calibration pipeline. We believe that exposing this previously overlooked attack surface, enabling independent validation, and motivating defenses provide substantial safety benefits that outweigh the remaining risk. We therefore decided to conduct and publish the research with the safeguards described above.

\cleardoublepage

\bibliographystyle{plain}
\bibliography{main}

\appendix

\section{Physical Realizability}
\label{sec:physical-realizability}

\paragraph{What the attacker has to do in the physical world.}
The artifact is a printed planar poster mounted on a rigid frame.
Its geometry $G_{\mathcal{A}}$ is fixed to dimensions consistent with a roadside billboard or construction sign, its pose $T_{\mathcal{A}}$ is restricted to ground-supported placements within a bounded distance and yaw of the vehicle's path, and its texture $C_{\mathcal{A}}$ is the only learnable component.
This restricts the attacker to manipulations that can be physically realized with consumer printing and standard structural mounting; the attacker requires no electronics, sensors, or active hardware in the artifact itself.
The entire physical-world deployment of \name\ reduces to: print the optimized $C_\mathcal{A}$ at the target resolution, mount it on a rigid board of the trained dimensions, and place it on the ground within the trained pose envelope.
No active hardware, no synchronization with the AV, no proximity to the AV at install time.

\paragraph{Robustness budget assumed in optimization.}
The EOT loop explicitly trains $C_\mathcal{A}$ to survive small deployment-time disturbances: pose jitter of a few degrees of rotation and a few tens of centimeters of translation, applied jointly to the artifact at every gradient step.
This budget reflects realistic placement noise (the attacker is unlikely to plant the poster within millimetres of the trained pose), residual ego-motion within the AV's calibration capture window, and minor LiDAR re-sampling differences between training and deployment. We do not model lighting or weather in the optimization; lighting consistency between camera and LiDAR is preserved by the rendering model.
Printing fidelity, by contrast, is no longer an assumption. The optimized texture is fully saturated RGB, which no CMYK process can reproduce, so rather than model printing as an attenuation factor, we push the texture through a measured print chain --- 8-bit quantization, gamut compression, and total-ink limiting --- and re-evaluate both stages on the result. Both survive it, so a standard CMYK print suffices and no wide-gamut process is needed. The optimization is additionally constrained by three properties of the printed surface that turn out to be \emph{measured requirements} rather than preferences --- continuous tone, a matte diffuse finish, and an opaque rigid mounting --- each of which we quantify, together with the print chain itself, in Section~\ref{sec:eval-print}. Print \emph{resolution} is not among them: the trigger survives a $20$\,mm ink smear across a $3.6$\,m board.

\paragraph{The board must sit on the sensor axis.}
The pose envelope is considerably tighter than the ``roadside billboard'' framing suggests. The board plane must be close to perpendicular to the sensor forward axis, with its centre near LiDAR height at $6$\,m, and Stage~1 degrades sharply with lateral offset: our jointly optimized poster fires on $0.89$ of held-out frames on-axis, $0.18$ at $1$\,m lateral offset, and $0.06$ at $2$\,m --- the gray-board floor. The same collapse appears on the other detector we tested and across all six alternative geometries we tried, so it is a property of these detectors' response rather than of the shape family we probed.
This directly qualifies the deployment story. A genuine roadside billboard sits several metres off the sensor axis, where the effect is nil; the on-axis geometry is met only when the vehicle is heading directly at the board, as at a bend, a junction, or a T-intersection. Every KITTI window used in this work is straight-road, so that deployment case is untested.


\section{Print Specification}
\label{sec:eval-print}

Because the artifact is a physical object, we state what must actually be manufactured, and treat each requirement as a measured result rather than a preference.

The poster is $3.600 \times 1.800$\,m printed $1{:}1$ at $1$\,mm/px, from an $89 \times 90$ texel grid of $40.4 \times 20.0$\,mm texels.
It must be \textbf{continuous-tone}: a smooth print gives $+14.79^\circ$ and $0.96$ fire, a hard-patch print reproducing the visible texel squares only $+10.38^\circ$ and $0.90$.
It must be \textbf{matte and diffuse} --- no gloss and no retroreflective film. This is not aesthetic: a specular return pushed off-range gives $0.00$ fire and reads \emph{more} aligned than a clean scene, and retroreflective film is worse still because it returns strongly at the correct range and so raises the detector's consistency score. NIR-absorbing substrates also degrade the trigger.
It must be \textbf{opaque and rigid, mounted flat}, because occlusion is part of the mechanism: on the Stage-1 detector, occlusion alone accounts for $0.26$ of the $0.28$ gray-board baseline, while a board visible to the LiDAR that hides nothing scores $0.00$.
Print \emph{resolution}, by contrast, is not critical --- the attack survives a $20$\,mm smear and fails only at $40$\,mm, far outside what any large-format printer produces.

The texture is fully saturated RGB, which no CMYK process can reproduce, so we measured the print chain rather than assuming it. Under $8$-bit quantization the numbers are unchanged ($0.89$ fire, $+14.85^\circ$); under an aggressive gamut compression toward luma the trigger falls to $0.75$ --- still above a useful threshold --- while Stage-2 damage slightly \emph{rises} to $+15.65^\circ$; under a realistic ink limit both hold ($0.85$, $+15.12^\circ$). Both stages therefore survive an ordinary CMYK print, and no wide-gamut process is required.

Placement is as trained: board plane perpendicular to the sensor's forward axis, centre at LiDAR height, $6.0$\,m ahead and on-axis, subject to the lateral tolerance noted above.

\end{document}